\documentclass{article}

\usepackage{PRIMEarxiv}
\usepackage{xcolor}
\usepackage[utf8]{inputenc} 
\usepackage[T1]{fontenc}    
\usepackage{hyperref}       
\usepackage{url}            
\usepackage{booktabs}       
\usepackage{amsfonts}       
\usepackage{nicefrac}       
\usepackage{microtype}      
\usepackage{soul}
\usepackage{lipsum}
\usepackage{amsmath}
\usepackage{algorithm}
\usepackage{algpseudocode}
\usepackage{fancyhdr}       
\usepackage{graphicx}       
\usepackage{caption}
\usepackage{subcaption}
\usepackage{hhline}
\usepackage[size=\tiny,]{todonotes}
\graphicspath{{media/}}     
\algrenewcommand\algorithmicrequire{\textbf{Input:}}
\algrenewcommand\algorithmicensure{\textbf{Output:}}
\usepackage{xcolor}
\newif\ifcommentcoloring
\commentcoloringtrue

\title{Divide and conquer: Learning chaotic dynamical systems with multi-step penalty neural ordinary differential equations}

\author{
  Dibyajyoti Chakraborty\\
  Information Sciences and Technology, \\
  Pennsylvania State University, \\
  University Park, PA, USA\\
  \texttt{d.chakraborty@psu.edu}
   \And
Seung Whan Chung \\
  Center for Applied Scientific Computing, \\
  Lawrence Livermore National Laboratory. \\
  Livermore, CA, USA.\\
  \texttt{chung28@llnl.gov}
  \And
  Troy Arcomano\\
  Environmental Science Division\\
  Argonne National Laboratory\\
   Lemont, IL, USA.
  \And
  Romit Maulik\\
  Information Sciences and Technology \\
  Pennsylvania State University, \\
  University Park, PA, USA \\
  \texttt{rmaulik@psu.edu} \thanks{Also at Mathematics and Computer Science Division, Argonne National Laboratory, Lemont, IL, USA.}
  }

\begin{document}
\maketitle

\begin{abstract}
Forecasting high-dimensional dynamical systems is a fundamental challenge in various fields, such as geosciences and engineering. Neural Ordinary Differential Equations (NODEs), which combine the power of neural networks and numerical solvers, have emerged as a promising algorithm for forecasting complex nonlinear dynamical systems. However, classical techniques used for NODE training are ineffective for learning chaotic dynamical systems. In this work, we propose a novel NODE-training approach that allows for robust learning of chaotic dynamical systems. Our method addresses the challenges of non-convexity and exploding gradients associated with underlying chaotic dynamics.
Training data trajectories from such systems are split into multiple, non-overlapping time windows. In addition to the deviation from the training data, the optimization loss term further penalizes the discontinuities of the predicted trajectory between the time windows. The window size is selected based on the fastest Lyapunov time scale of the system. Multi-step penalty(MP) method is first demonstrated on Lorenz equation, to illustrate how it improves the loss landscape and thereby accelerates the optimization convergence. MP method can optimize chaotic systems in a manner similar to least-squares shadowing with significantly lower computational costs. Our proposed algorithm, denoted the Multistep Penalty NODE, is applied to chaotic systems such as the Kuramoto--Sivashinsky equation, the two-dimensional Kolmogorov flow, and ERA5 reanalysis data for the atmosphere. It is observed that MP-NODE provide viable performance for such chaotic systems, not only for short-term trajectory predictions but also for invariant statistics that are hallmarks of the chaotic nature of these dynamics.
\end{abstract}

\keywords{Neural ordinary differential equations \and Chaotic dynamical systems \and Scientific machine learning.}

\section{Introduction}

Dynamical systems are ubiquitous with examples such as weather, fluid flows, chemical reactions, etc. Ordinary differential equations are commonly used to describe the time evolution of deterministic dynamical systems. Many common dynamical systems show chaotic behavior (for example the behavior of the atmosphere), where the time evolution of their trajectories is extremely sensitive to initial conditions. For such systems, small perturbations in initial conditions are amplified leading to significant deviations over time. It must be noted that these deviations are physical and do not represent errors in the model -- also commonly identified as the hallmark of `deterministic chaos'. The data-driven modeling of chaotic systems is thus complicated by the lack of a simple error metric to guide optimization and has engendered many unique modeling strategies.
We remark that ``learning'' of a chaotic system is defined not simply by accurately predicting instantaneous trajectory over time, but more importantly by matching its invariant statistics at the attractors.



Neural Ordinary Differential Equations~(NODEs)~\cite{chen2018neural} combine the universal approximation capability of neural networks with the rich mathematical resources and efficiency of numerical differential equation solvers. NODE shows particular promise in learning dynamical systems where classical theory-driven models may be insufficiently accurate or computationally expensive for the modeling of phenomena. Since NODE offers a continuous time evolution of dynamics, they can also be used for time series modeling with irregularly spaced data. Even though NODE has been explored extensively in the literature over a wide range of research areas~\cite{dang2023constrained,hwang2021climate,luo2022stability}, there are open questions regarding NODE: their efficient parallelization~\cite{chen2018neural}; how they may learn stiff systems~\cite{kim2021stiff}; and what classes of functions they struggle to approximate~\cite{dupont2019augmented}.

Chaotic systems pose a particular challenge to NODE. Training NODE (or any other neural network) typically involves gradient-based optimization for a loss term defined as the deviation from ground truth. Such gradient-based optimization is notoriously challenging for chaotic dynamical systems. The recursive application of horseshoe mapping makes the chaotic system extremely sensitive to small perturbations.
Such extreme sensitivity manifests as the exploding gradient in optimization~\cite{chung2022optimization}. Furthermore, the loss term under chaotic systems becomes extremely non-convex in parameter space, where the gradient often guides the optimization toward poor local minima.


This challenge has connections with the exploding/vanishing gradient problem in deep learning~\cite{pascanu2013difficulty, hanin2018neural, philipp2017exploding}. Just like for evolving chaotic dynamics, the recursive application of multiple, deep neuron layers either explodes or vanishes the gradient, hindering the training.
In the machine learning community, this exploding gradient problem has been essentially bypassed by changing the structure of the neural networks: keeping the shallow residual networks~\cite{philipp2018gradients}; using echo state networks (ESN)~\cite{sun2020review}; and regularizing the dynamics of the network~\cite{ven2021regularization}. For applications of NODE to chaotic dynamics, such options are not viable, since the \emph{emergence of chaos in trained model predictions is a desirable property}.

For learning chaotic dynamics, it is therefore common to focus on ergodic quantities instead of instantaneous trajectories. Several algorithms have been proposed for computing the gradients of ergodic quantities in chaotic dynamics: ensemble averaging~\cite{lea2002sensitivity}; linear response theory~\cite{eyink2004ruelle}; a probability density function-based method~\cite{thuburn2005climate}; cumulant-truncation~\cite{craske2019adjoint}; and least-squares shadowing~\cite{blonigan2013new, ni2017sensitivity, chandramoorthy2017analysis, ni2019adjoint, ni2023fast}.

However, these ergodic sensitivity methods often suffer either from poor convergence due to the exploding gradient or from extensive computational costs~\cite{lea2002sensitivity,eyink2004ruelle,thuburn2005climate,craske2019adjoint}.
The Least-Squares Shadowing(LSS) approach provides accurate gradients based on the shadowing lemma, which has been proven effective for various chaotic systems~\cite{blonigan2013new, ni2017sensitivity, chandramoorthy2017analysis, ni2019adjoint, ni2023fast}. However, this involves solving an optimization problem for the gradient at each training step, which in turn significantly increases the computational cost of the training. This will be further examined in Section~\ref{sensitivity analysis}.


Recently, the multi-step penalty method~\cite{chung2022optimization} has been proposed for the optimization of chaotic turbulent flows.
This optimization method divides the time domain into multiple intermediate steps, where the discontinuities in time are penalized throughout the optimization.
This effectively bypasses the difficulty of computing gradients for chaotic systems at the cost of typical gradient computation. The reader may also note similarities between this method to classical techniques such as the multiple shooting method~\cite{van1975nonlinear,bock1981numerical,bock1984multiple}.
In this paper, we propose an algorithm that can improve the training of NODE for chaotic systems, as an extension of the multi-step penalty method~\cite{chung2022optimization} to NODE framework -- the Multistep Penalty Neural Ordinary Differential Equation (MP-NODE). Specifically, this novel algorithm allows back-propagation of the neural networks gradient through long rollouts of chaotic systems, which otherwise face problems of exploding gradients~\cite{mikhaeil2022difficulty}. The proposed method is tested on several representative chaotic systems: the Lorenz-63 equations; the Kuramoto--Sivashinsky equations; and a two-dimensional forced turbulence problem given by the Kolmogorov Flow. The trained NODE not only provides accurate predictions on short-term trajectories but also provides long-term adherence to invariant statistics, thereby indicating that the true physics of the system has been captured. This paves the way for building NODE for realistic chaotic forecasting applications in various scientific domains.

The rest of the paper is structured as follows:
the \hyperref[lit]{Background} section explores the various modifications and applications of NODE and the previous works in the data-driven prediction of chaotic systems. It also highlights the causes of the problems in sensitivity analysis of chaotic systems and the LSS technique. The \hyperref[meth]{Methodology section} focuses on the mathematical formulations of the proposed MP-NODE method. The \hyperref[results]{Results section} explores its performance for predictions of several representative chaotic systems. Finally, the \hyperref[d and c]{Discussion} section summarizes the contributions, impact, limitations, and future opportunities of our work.

\section{Background}\label{lit}
\subsection{Neural Ordinary Differential Equations}

A Neural Ordinary Differential Equation~\cite{chen2018neural} is defined as 
\begin{equation}\label{NODE}
\begin{aligned}
    &\frac{d\textbf{q}(t)}{dt} = \mathcal{R}(t,\textbf{q}(t),\boldsymbol{\Theta})\\
    &\textbf{q}(0) = \textbf{q}_0,
\end{aligned}
\end{equation}
where $\mathcal{R}:\mathbb{R}\times\mathbb{R}^{d_1 \times d_2 \dots \times d_n} \times \mathbb{R}^{d_{\boldsymbol{\Theta}}} \longrightarrow \mathbb{R}^{d_1\times d_2 \dots \times d_n}$ is generally a simple feed-forward network, $\boldsymbol{\Theta}$ is a vector of learnable parameters, and $\textbf{q}:\mathbb{R}\longrightarrow\mathbb{R}^{d_1\times d_2 \dots \times d_n}$ is the solution to the differential equation. NODE are considered as the continuous limit of Residual Networks (ResNets)~\cite{he2016deep}.
Since the evolution of hidden states in a Resnet is similar to the Euler approximation of an ODE, the advantage of NODE is to use more efficient and accurate ODE solvers~\cite{alexander1990solving}.
Secondly, most advanced deep learning techniques like GRUs~\cite{cho2014learning} and LSTMs~\cite{hochreiter1997long} use autoregressive procedures that are very similar to discretized differential equations.
Another advantage to using NODE is the ability to use the adjoint sensitivity method which can compute gradients with far-reduced memory requirements~\cite{boltyanskiy1962mathematical}.
NODE can achieve the same accuracy as ResNet~\cite{he2016identity} for the MNIST~\cite{lecun1998mnist} dataset with one-third number of parameters due to its low memory cost \cite{chen2018neural} (i.e., it uses a fixed amount of memory that is constant with the depth of neural network). If memory is not a bottleneck, current automatic differentiation libraries like Jax~\cite{jax2018github} can efficiently use backpropagation to compute gradients for training NODE.
Regardless of the mode of gradient computation, vanilla NODE are unable to learn chaotic dynamics \cite{linot2023stabilized} if trained using a simple mean-squared-error minimization. This motivates our algorithmic development described in the following sections.

While interest in Neural Differential Equations has grown recently, its roots lie in substantially old literature. Chu et al.(1991) used artificial neural networks (ANNs) to predict the dynamics of a continuous time-evolving system~\cite{chu1991neural}. Rico et al. (1992) used ANNs to predict short-term dynamics of experimental chemical reaction data~\cite{rico1992discrete}. To the best of our knowledge, they were the first to use and state the limitations of NODE in predicting dynamical systems. They also used NODE to predict latent dynamics \cite{rico1994continuous} (i.e., dynamics in a reduced-dimensional space) as an extension to their previous work and further solved stiff ODE systems as well \cite{rico1993continuous}.

Recently, NODE was popularized in the era of data-intensive machine learning by the work of Chen et al. (2018) \cite{chen2018neural}. We denote their formulation as the 'vanilla' NODE. Various improvements and modifications have been proposed for the vanilla NODE. Since NODE preserves the topology of the input space, which limits the approximation of a family of functions, Dupont et al.(2019)~\cite{dupont2019augmented} developed an augmented NODE that is more stable and generalizable. They proposed to augment the input space that can avoid intersections of ODE trajectories by providing additional dimensions to lift the points. Norcliffe et al.(2020)~\cite{norcliffe2020second} went beyond the augmented NODE to explicitly incorporate the second-order behavior of dynamical systems. The vanilla NODE just evolves the dynamics based on initial conditions and has no mechanism to adjust its trajectory based on subsequent observations. Neural Controlled Differential Equations (NCDE) tackles this issue by encompassing incoming data using Riemann--Stieltjes integral in the training of a NODE to improve its performance for irregular time series \cite{kidger2020neural}. NCDEs can be interpreted as an approximation for Recurrent Neural Networks (RNNs) with infinite depth. Ghosh et al. (2020) proposed the temporal regularization of NODE for improving performance and reducing training time~\cite{ghosh2020steer}. Kim et al. (2021) showed that modified ODE solvers like Implicit-Explicit methods can boost the capability of NODE to solve stiff dynamical systems~\cite{kim2021stiff}. In our past work, Linot et al. (2023) added a stability-promoting linear term to the right-hand side of Equation \ref{NODE}, which improved its capability to predict the long-term dynamics of chaotic systems~\cite{linot2023stabilized}. Yildiz et al. (2019) presented a generative model based on NODE using Variational Auto-Encoders\cite{yildiz2019ode2vae}. Zhang et al. (2019) used a coupled ODE architecture for evolving the weight and activation functions of a NODE~\cite{zhang2019anodev2}. They demonstrated significant gains in accuracy with the same memory and computational time cost. Poli et al. (2019) used a graph neural network to represent the function on the right hand side of a NODE~\cite{poli2019graph}. Liu et al. (2019)~\cite{liu2019neural} used stochastic noise to stabilize NODE. Massaroli et al.~(2020) used orthogonal basis functions to represent the parameters of a NODE and improved its interpretability and generalizability~\cite{massaroli2020dissecting}.


 NODE has been heavily studied for applications to predicting various dynamical systems. Maulik et al. (2020) compared NODE and long short-term memory (LSTM) for predicting the viscous burgers equation\cite{maulik2020time}. It has also been used to perform the time series modeling of turbulent kinetic energy \cite{portwood2019turbulence}, stochastic differential equations~\cite{tzen2019neural}, reduced order modeling for dynamics of the flow past a cylinder~\cite{rojas2021reduced}, modeling the chaotic dynamics in Kuramoto-Sivashinsky equations~\cite{linot2022data}, prediction of battery degradation~\cite{bills2020universal}, climate modeling~\cite{hwang2021climate}, continuous emotion prediction~\cite{dang2023constrained}, link prediction in graphs~\cite{luo2023graph}, improving the adversarial robustness of image classifiers~\cite{carrara2022improving}, efficient solving of chemical kinetics~\cite{owoyele2020chemnode}, and chaotic complex Ginsburg Landau equation~\cite{gelbrecht2021neural}. Among these studies, we specifically point out the framework of the multiple-shooting NODE~\cite{turan2021multiple} which uses the concept of introducing local discontinuities in the training data trajectories. This is found to be beneficial for purposes such as training with irregularly sampled trajectories~\cite{iakovlev2022latent}, improving efficiency and scalability~\cite{lagemann2023invariance}, learning from sparse observations~\cite{jordana2021learning}, and forecasting with limited training data~\cite{abrevaya2023effective}. However, the connection to learning the invariant statistics of chaotic dynamics has not been explored.

\subsection{Stable Data-Driven Prediction of Chaotic Systems}

Learning long term stable dynamics of chaotic systems has been a challenge for data-driven modeling of dynamical systems. It is well-known that capturing long term dynamics through gradient descent is difficult~\cite{bengio1994learning,schmidt2019identifying}. Standard time-series learning methods include recurrent neural networks (RNN)~\cite{lipton2015critical}. RNNs suffer from vanishing or exploding gradients~\cite{pascanu2013difficulty} which motivated the development of LSTM~\cite{hochreiter1997long}, gated recurrent units (GRU)\cite{cho2014learning}, and other derivative methods that enforce constraints on the recurrence matrix of the RNN~\cite{chang2019antisymmetricrnn,arjovsky2016unitary,jing2019gated}. Other techniques popularly used for the prediction of chaotic systems are reservoir computing~\cite{pathak2018model}, teacher forcing~\cite{williams1989learning}, and multiple shooting~\cite{voss2004nonlinear}. Recent developments have used loss functions based on invariant statistics for stable prediction of chaotic systems~\cite{schiff2024dyslim}. However, these tend to rely on matching empirical statistics of predicted trajectories which can place significant memory and computational burdens during optimization. In this work, we intentionally avoid loss-function based penalties for learning invariant statistics. An interesting observation in all of the above works is that autoregressive models tend to overfit on short term dynamics and diverge in longer rollouts. A solution to this is to use longer rollouts in training~\cite{keisler2022forecasting}. However, the gradients become less useful when backpropagating through longer rollouts as it is theoretically proven that gradients of RNN-type methods always diverge for chaotic systems~\cite{mikhaeil2022difficulty}. We claim that the solution to training data-driven models for learning chaotic dynamics may be found by following in the footsteps of rich literature devoted to sensitivity analysis for chaotic systems, which is introduced subsequently


\subsection{Sensitivity Analysis for Chaotic systems}\label{sensitivity analysis}
Standard NODEs are effective for approximating non-chaotic dynamical systems. However, standard gradient-based techniques to train NODEs are certain to lead to poor learning, since gradient computation in chaotic systems is ill-posed. Evidence for this was provided in Chung et al.(2022) \cite{chung2022optimization} where it was demonstrated that standard gradient-based optimization fails for several chaotic systems. To further explain the specifics of this research, we start by defining a standard ODE-constrained optimization problem defined as
\begin{equation}\label{ODE optim}
    \mathrm{minimize}\ \mathcal{J}[\mathbf{q},\boldsymbol{\Theta}]\in\mathbb{R}\quad\mathrm{such~that}\quad\mathcal{N}[\mathbf{q};\boldsymbol{\Theta}]=\mathbf{0},
\end{equation}
where $\mathbf{q}$ and $\mathbf{\Theta}$ are constrained by an ODE, 
\begin{equation}\label{ODE}
    \mathcal{N}[\mathbf{q};\boldsymbol{\Theta}] \equiv \frac{d\mathbf{q}}{dt}-\mathcal{R}[\mathbf{q};\boldsymbol{\Theta}]=\mathbf{0},
\end{equation}
with initial condition
\begin{equation}
    \mathbf{q}(t_i)=\mathbf{q_i}.
\end{equation}
An objective function to be minimized is typically defined as
\begin{equation}
    \mathcal{J}[\mathbf{q},\boldsymbol{\Theta}]=\frac{1}{t_f-t_i}\int_{t_{i}}^{t_{f}}\mathcal{I}[\mathbf{q}(t),\boldsymbol{\Theta}(t)]\,dt.
\end{equation}
To compute the sensitivity $\frac{d\mathcal{J}}{d\boldsymbol{\Theta}}$, a conventional tangent(or forward) method\cite{bryson2018applied} of sensitivity analysis can be used. Using $\textbf{v}(t;\boldsymbol{\Theta})=\frac{\partial \textbf{q}(t;\boldsymbol{\Theta})}{\partial \boldsymbol{\Theta}}$ equation \ref{ODE} is converted to its linearized form (tangent equation).
\begin{equation}\frac{d\textbf{v}}{dt}=\frac{\partial \mathcal{R}(\mathbf{q};\boldsymbol{\Theta})}{\partial \textbf{q}}\textbf{v}+\frac{\partial \mathcal{R}(\mathbf{q};\boldsymbol{\Theta})}{\partial \boldsymbol{\Theta}},\end{equation}
with initial condition
\begin{equation}
\textbf{v}|_{t=t_i}=\frac{d\textbf{q}_i}{d\boldsymbol{\Theta}}.
\end{equation}
The solution to the tangent equation, can be used to obtain the gradient of the objective,
\begin{equation}\label{Sensitivity_eqn}
\frac{d\mathcal{J}}{d\boldsymbol{\Theta}}=\frac{1}{t_f-t_i}\int_{t_i}^{t_f}\left(\frac{\partial \mathcal{I}(\textbf{q},\boldsymbol{\Theta})}{\partial \textbf{q}}\textbf{v}+ \frac{\partial \mathcal{I}(\textbf{q},\boldsymbol{\Theta})}{\partial \boldsymbol{\Theta}}\right)dt.
\end{equation}

For chaotic systems, the tangent variable $\textbf{v}(t;\boldsymbol{\Theta})$ in Equation \ref{Sensitivity_eqn} grows at an exponential rate (given by the Lyapunov Exponent), failing the standard method to compute a useful gradient for optimization. Ergodic quantities of the chaotic dynamics are proven to be differentiable, though the actual evaluation of their gradients is challenging~\cite{lea2002sensitivity,eyink2004ruelle,thuburn2005climate,craske2019adjoint,blonigan2013new}. 

A solution to the gradient computation problem exploits the Shadowing Lemma~\cite{katok1995introduction}, which states that for a reference solution of a chaotic dynamical system, there exists a shadow trajectory that is always close to the original trajectory for a small perturbation of the parameters of the dynamical system. The Least Square Shadowing(LSS) method\cite{wang2014least} obtains a shadowing trajectory $\textbf{q}$ with respect to the reference solution $\textbf{q}_r$, for an arbitrary initial condition and a time transformation $\tau(t) : (0,T) \longrightarrow \mathbb{R}$ , by optimizing
    \begin{equation}\label{LSS1}
    \begin{aligned}
    \begin{aligned}\operatorname*{minimize}_{\tau,\textbf{q}}\frac{1}{T} \int_{0}^{T}\left\|\textbf{q}(\tau(t)\right)-\textbf{q}_{r}(t)\|^2dt,\end{aligned} \\
    \text{such\ that\ }
    \frac{d\textbf{q}}{d\tau}=\mathcal{R}\left(\textbf{q},\boldsymbol{\Theta}+d\boldsymbol{\Theta}\right).
    \end{aligned}
\end{equation}
A modified version of Equation \ref{LSS1} which has a better condition number on solving is
\begin{equation}\label{LSS2}
    \begin{aligned}
    \operatorname*{minimize}_{\tau,\textbf{q}}\frac{1}{T}&\int_{0}^{T}\left(\left\|\textbf{q}(\tau(t)\right)-\textbf{q}_{r}(t)\right\|^{2}+\alpha^{2}\left(\frac{d\tau}{dt}-1\right)^{2}dt,\\
    & \text{such\ that\ } \frac{d\textbf{q}}{d\tau}=\mathcal{R}\left(\textbf{q},\boldsymbol{\Theta}+d\boldsymbol{\Theta}\right).
    \end{aligned}
\end{equation}
The tangent form of Equation \ref{LSS2} is given as
\begin{equation}\label{LSS3}
    \begin{aligned}&\operatorname*{minimize}_{\eta,\textbf{v}}\frac{1}{T}\int_0^T(\|\textbf{v}\|^2+\alpha^2\eta^2)dt,\\
    & \text{such\ that\ }\frac{d\textbf{v}}{dt}=\frac{\partial f}{\partial \textbf{q}}\textbf{v}+\frac{\partial \mathcal{R}}{\partial \boldsymbol{\Theta}}+\eta \mathcal{R}(\textbf{q}_r,\boldsymbol{\Theta}),\end{aligned}
\end{equation}
which determines the tangent variables for $\mathbf{q}$ and $\frac{d\tau}{dt}$, respectively,
\begin{equation}
    \textbf{v}(t)=\frac d{d\boldsymbol{\Theta}}(\textbf{q}(\tau(t;s);\boldsymbol{\Theta})),\quad\eta(t)=\frac {d}{d\boldsymbol{\Theta}}\frac{d\tau(t;\boldsymbol{\Theta})}{dt}.
\end{equation}
The sensitivity of the objective can be computed using
    \begin{equation}\frac{d \mathcal{J}}{d\boldsymbol{\Theta}}\approx\frac{1}T\int_0^T(\frac{\partial \mathcal{I}}{\partial \textbf{q}}\textbf{v}+\frac{\partial \mathcal{I}}{\partial \boldsymbol{\Theta}}+\eta(\mathcal{I}-\overline{J}))dt,\quad\mathrm{where~}\overline{J}=\frac{1}T\int_0^T\mathcal{I}dt .\end{equation}
The optimization problem \ref{LSS3} can be solved by formulating the Lagrangian,
\begin{equation}
    \begin{aligned}L&=\int\limits_0^T\left(\textbf{v}^\top \textbf{v}+\alpha^2\eta^2+2w^\top\left(\frac{d\textbf{v}}{dt}-\frac{\partial \mathcal{R}}{\partial \textbf{q}}\textbf{v}-\frac{\partial \mathcal{R}}{\partial \boldsymbol{\Theta}}-\eta \mathcal{R}\right)\right)dt\end{aligned}
\end{equation}
Setting the first order derivative of the Lagrangian to zero, the following Karush--Kuhn--Tucker system is obtained.
\begin{equation}
   \begin{aligned}&\frac{d\textbf{v}}{dt}-\frac{\partial \mathcal{R}}{\partial \textbf{q}}\textbf{v}-\frac{\partial \mathcal{R}}{\partial \boldsymbol{\Theta}}-\eta \mathcal{R}=0\\&\frac{dw}{dt}+\frac{\partial \mathcal{R}}{\partial \textbf{q}}^\top w-\textbf{v}=0\\&w(0)=w(T)=0\\&\alpha^2\eta-w^\top \mathcal{R}=0.\end{aligned}
\end{equation}
This can be solved numerically by converting the system of ODE to a system of linear equations with a computational cost of $O(mn^3)$ to get the sensitivities, where $m$ is the number of time steps, and $n$ is the degrees of freedom of the dynamical system, which is the dimension of state variable $\textbf{q}$ multiplied to the dimension of parameters $\boldsymbol{\Theta}$ \cite{blonigan2016least}. LSS has been used to optimize several chaotic systems including turbulent flows~\cite{blonigan2014least} and flow around a two-dimensional airfoil~\cite{blonigan2018least}. However, this is computationally expensive when performed in each iteration of an optimization, especially for deep learning algorithms where there are millions of trainable parameters and potentially several thousands of gradient computations for a prolonged optimization.
 
Recently, several improvements to vanilla LSS have been proposed~\cite{chandramoorthy2017analysis,ni2023fast,chandramoorthy2017analysis,ni2017sensitivity}. Ni et al. (2019) improved upon standard adjoint sensitivity and proposed the Non-Intrusive Least Square Adjoint Shadowing (NILSAS) technique for chaotic dynamical systems~\cite{ni2019adjoint}. Chandramoorthy et al. (2020) proposed a procedure to use automatic differentiation for computing shadowing sensitivity~\cite{chandramoorthy2020variational}. However, an efficient method with low computational and memory requirements for obtaining sensitivity of chaotic systems is still an active area of research.

\section{Methodology}\label{meth} 






\subsection{Multistep Penalty Method}

To address the issues with standard gradient-based optimization for chaotic dynamical systems we turn to a multi-step penalty method (MP) proposed by Chung et al. (2022)~\cite{chung2022optimization}. In this work, the authors proposed the introduction of intermediate initial conditions ($\mathbf{q_k}^+$) such that for $t_k = t_i + kT$,
\begin{equation}\label{MP}
	\begin{aligned}
		&\frac{d\mathbf{q}}{dt}-\mathcal{R}[\mathbf{q};\boldsymbol{\Theta}]=\mathbf{0}\quad\mathrm{~for~}t\in[t_k,t_{k+1})\\
		&\mathbf{q}(t_k)=\mathbf{q}_k^+\quad\mathrm{~for~}k=\mathbf{0},\ldots,N-1.
	\end{aligned}
\end{equation} 
The time domain is split based on the Lyapunov time of the system. This prevents the exploding gradient and the extremely non-convex objective functional. The intermediate constraints are additionally imposed,
\begin{equation}
	\Delta\mathbf{q}_k\equiv\mathbf{q}_k^+-\mathbf{q}(t_k^-)=0\quad\mathrm{~for~}k=1,\ldots,N-1,
\end{equation}
where,
\begin{equation}
	\mathbf{q}(t_k^-)=\mathbf{q}_{k-1}^++\int_{t_{k-1}}^{t_k^-}\mathcal{R}[\mathbf{q};\mathbf{\Theta}]dt.
\end{equation}
Now, a penalty loss $\mathcal{P}$, which penalizes the intermediate constraints with a regularizing constant $\mu$, is added to the objective to form an augmented objective function $\mathcal{J}_A$ 
\begin{equation}\label{MP}
	\mathcal{J}_A[\mathbf{q},\mathbf{\Theta};\{\mathbf{q}_k^+\},\mu]=\mathcal{J}[\mathbf{q},\Theta]+\mathcal{P}[\{\Delta\mathbf{q}_k\},\mu],
\end{equation}
With $\{\mathbf{q}_k^+\}=(\mathbf{q}_1^+,\mathbf{q}_2^+,\ldots,\mathbf{q}_{N-1}^+)\text{ and }\{\Delta\mathbf{q}_k\}=(\Delta\mathbf{q}_1,\Delta\mathbf{q}_2,\ldots,\Delta\mathbf{q}_{N-1}) $, the subproblem is formulated as,
\begin{equation}\label{penalty_based_optimization}
	(_i\{\mathbf{q}_k^+\},_i\boldsymbol{\Theta})=\operatorname{argmin}_{\{\mathbf{q}_k^+\},\boldsymbol{\Theta}}J_A[\mathbf{q},\boldsymbol{\Theta};\{\mathbf{q}_k^+\},_i\mu],
\end{equation}
Adjoint sensitivity or automatic differentiation can be used to compute the gradients of Equation \ref{MP}. The MP method can solve the problems of exploding gradients in chaotic optimization as shown further in the paper. The computational cost of this method is of the same order as the cost of computing gradients in the classical way using automatic differentiation.

\subsection{Multi-step Penalty Neural Ordinary Differential Equations}

\begin{figure}[h]
    \centering
    \includegraphics[width=\textwidth]{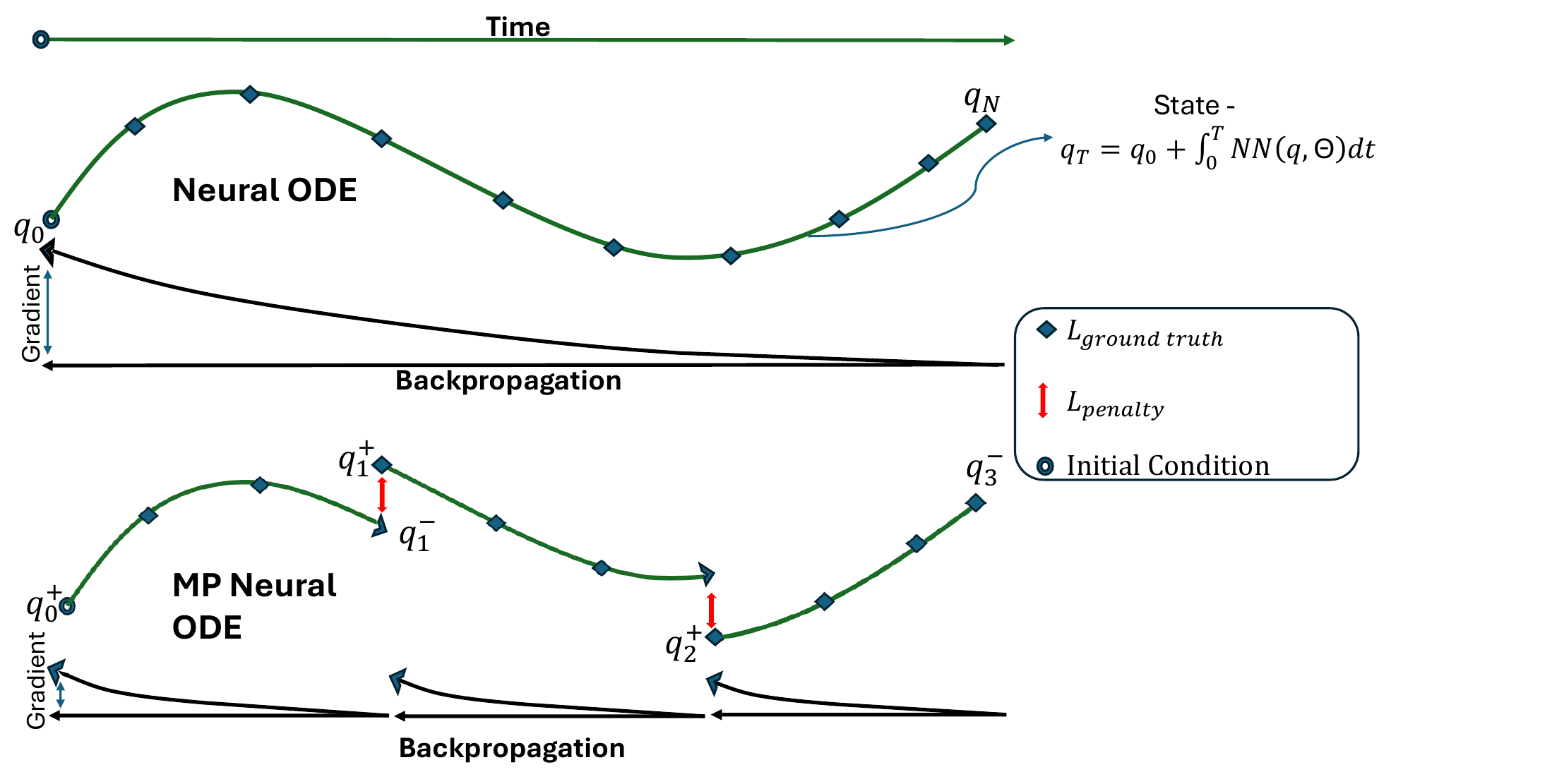}
    \caption{A schematic for multistep penalty based optimization of a data-driven dynamical systems. Discontinuities are introduced into the state evolution as learnable parameters (denoted the `penalty' term in the loss) during the process of optimization. Over the course of the optimization, the penalty term is driven to zero.}
    \label{fig:MP_schema}
\end{figure}

The requirement of an efficient algorithm for backpropagating through long rollouts is necessary to learn invariant statistics of chaotic systems. Consequently, we propose a multistep penalty optimization to train NODE which has a similar computational cost as vanilla NODE. A schematic for comparison between a vanilla NODE and MP-NODE is shown in figure \ref{fig:MP_schema}. The multistep penalty optimization can be extended to NODE by using a neural network as the RHS of Equation \ref{MP} and changing the loss function $\mathcal{J}$ to a Mean Squared Error (MSE) loss between the prediction and the ground truth.
We propose to use the Multi-step Penalty Neural ODE (MP-NODE) defined as,
\begin{equation}
\begin{aligned}
		&\frac{d\textbf{q}(t)}{dt} - \mathcal{R}(\textbf{q}(t),t,\boldsymbol{\Theta}) = 0          \quad\mathrm{~for~}t\in[t_k,t_{k+1})\\
		&\textbf{q}(t_k)=\textbf{q}_k^+\quad\mathrm{~for~}k=0,\ldots,n-1.
	\end{aligned}
\end{equation}
The loss function can be augmented with the penalty loss as

\begin{equation}\label{MP loss equation}
    \mathcal{L} = \mathcal{L}_{GT} + \frac{\mu}{2}\ \mathcal{L}_{P}
\end{equation}
where,
\begin{equation}
   \mathcal{L}_{GT} = \frac{\sum_{i=1}^{i=N}|\textbf{q}_i - \textbf{q}_i^{true}|^2 }{2N}; \quad \mathcal{L}_{P} = \frac{\sum_{k=1}^{k=n-1}|\textbf{q}_k^+ - \textbf{q}_k^-|^2 }{n-1},
\end{equation}
are the loss with respect to ground truth and penalty loss with respect to discontinuity respectively. For k = 1,2,...n
\begin{equation}
    \textbf{q}_k^- = \textbf{q}_{k-1} + \int_{t_{k-1}^+}^{t_k^-} \mathcal{R}(\textbf{q}(t),t,\boldsymbol{\Theta})\,dt.
\end{equation}
With strategically varying the penalty strength $\mu$ one can accommodate for local discontinuities in $z$ (given by $|\textbf{q}_k^+ - \textbf{q}_k^-|$) to effectively reduce $\mathcal{L}_{GT}$ in initial iterations and then increase $\mu$ gradually to reduce $\mathcal{L}_{P}$. The rule of thumb for penalty strength $\mu$ is to start with a low value(for example $10^{-6}$ in Section \ref{KS_problem}) and increase it once the the gradient reaches a certain cutoff\cite{chung2022optimization,bertsekas2014constrained}. In our experiments, the cutoff point for updating $\mu$ is determined with respect to the number of iterations chosen empirically observing the loss curve. We present three different algorithms in appendix \ref{Algo_Sec} for the multistep penalty optimization for NODE. The best performing of them as per our experiments is introduced in algorithm \ref{Algo1}.
\begin{algorithm}[h]
\caption{Multi-step NeuralODE - Formulation 1}\label{Algo1}
\begin{algorithmic}
\For{$\mu$ in  \{$\mu_i$\}} 
    \While{ $j<$  n\_batches}
    \State get\_batch($t$, $\textbf{q}$, ..) \Comment{Get a new batch}
    \State Initialize \textbf{$q_k$} \Comment{Initialize discontinuities for the new batch}
    \State Params $\gets$($\boldsymbol{\Theta}$, \textbf{$q_k$}) \Comment{Append neural network parameters and \textbf{$q_k$} as set of trainable parameters}
    \State ($\nabla \mathcal{L}_{\boldsymbol{\Theta}}, \nabla \mathcal{L}_{\boldsymbol{q_k}}$) = compute\_gradients($\mathcal{L}$,Params,$\mu_i$) \Comment{Compute gradients using Adjoint or autograd}
    \State Params $\gets$ optimizer.step(Params,$\nabla \mathcal{L}_{\boldsymbol{\Theta}}, \nabla \mathcal{L}_{\boldsymbol{q_k}}$) \Comment{Update the trainable parameters} 
    \EndWhile
\State update $\mu$ \Comment{Increase Penalty Strength}
\EndFor
\end{algorithmic}
\end{algorithm}

\section{Results}\label{results}

In section~\ref{subsec:lorenz}, we first demonstrate the effectiveness of Multistep Penalty(MP) optimization for chaotic dynamics in the aspects of exploding gradients and non-convexity of loss landscape.
The proposed MP NODE method is demonstrated subsequently in sections \ref{KS_problem} through \ref{subsec:era5}.

\subsection{Effectiveness of Multistep Penalty method}\label{subsec:lorenz}

\subsubsection{Exploding Gradients}
We take the same problem as mentioned in the original LSS~\cite{wang2014least} paper for a proof of concept demonstration of the proposed method. It introduces a controlled Lorenz-63 system,
\begin{equation}
    \begin{aligned}
        &\frac{dx}{dt}=\sigma(y-x)\\
        &\frac{dy}{dt}=x(\rho-z)-y\\
        &\frac{dz}{dt}=xy-\beta z,
    \end{aligned}
\end{equation}
where $\rho$ is the control parameter. The objective function for control is given by 
\begin{equation}\label{eq:lorenz-J}
    J = \lim_{T\to \infty}\frac{1}{T}\int_{0}^{T}z\,dt.
\end{equation}
\begin{figure}[hp]
    \centering
    \includegraphics[width=0.99\textwidth]{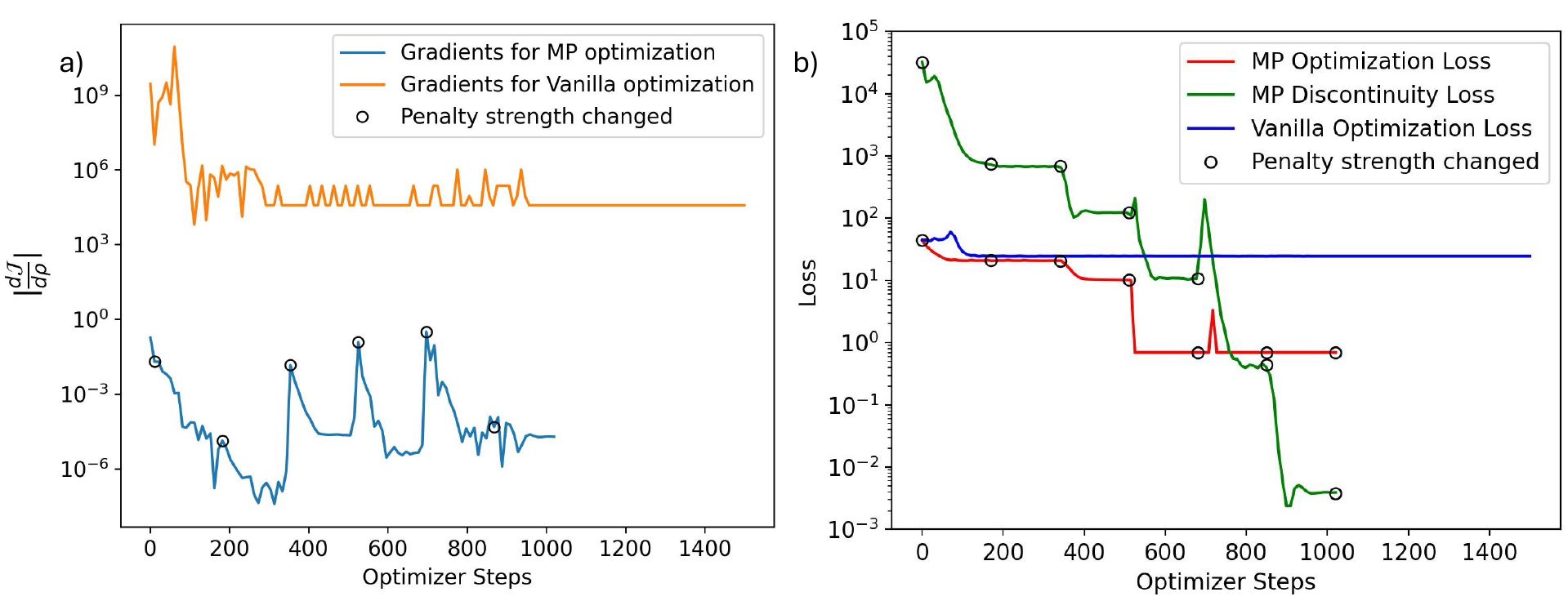}
    \caption{a) A comparison between gradients of the objective function with respect to parameter $\rho$ with increasing optimization steps for vanilla optimization (i.e., without any multistep penalties) and MP optimization. The black circle represents the locations where the penalty strength is increased.b) Objective values for MP optimization and vanilla optimization with optimization steps. The MP optimization reaches the theoretical minimum(approx 0.694) whereas the vanilla optimization reaches a plateau much higher than the minima. }
    \label{fig:lorenz}
\end{figure}
To convert the original problem to a minimization problem without reducing its complexity, the $z$ in the integral is changed to $|z|$ and we approximate $J$ by taking a sufficiently long $T=20$ without loss of generality. A standard calculation of the gradient of $J$ (using finite differences or automatic differentiation) leads to very large magnitudes ($O(10^{100})$) which are unsuitable for learning. Figure \ref{fig:lorenz} shows that gradients computed using standard automatic differentiation explode to $O(10^8)$ at the start of optimization and later get to a constant of approximately $O(10^3)$ which is not useful to reduce the objective function.
In contrast, multi-step penalty gradients are significantly improved and successfully reduce the objective function as shown in figure \ref{fig:lorenz}.
Note that the loss and gradients have periodic jumps at points where the penalty strength $\mu$ is increased. In this problem, we started with a low penalty strength of $\mu = 10^{-5}$ and increased it by a multiple of 10 after every 170 optimizer steps based on when the gradient shown in Figure \ref{fig:lorenz}(a) starts to plateau. These values are chosen empirically and can be tuned further for faster optimization. This proof-of-concept demonstration shows that the issue of exploding gradients for chaotic systems can be solved by MP optimization. Furthermore, the cost of MP optimization is $O(m+n)$ where $m$ is the number of time-steps and $n$ is the degree of freedom (sum of the number of trainable parameters and the dimension) which is significantly lower than the cubic cost of LSS.
\par
\subsubsection{Non-Convexity of Loss Landscape}
\begin{figure}[h]
    \centering
    \includegraphics[width=0.9\textwidth]{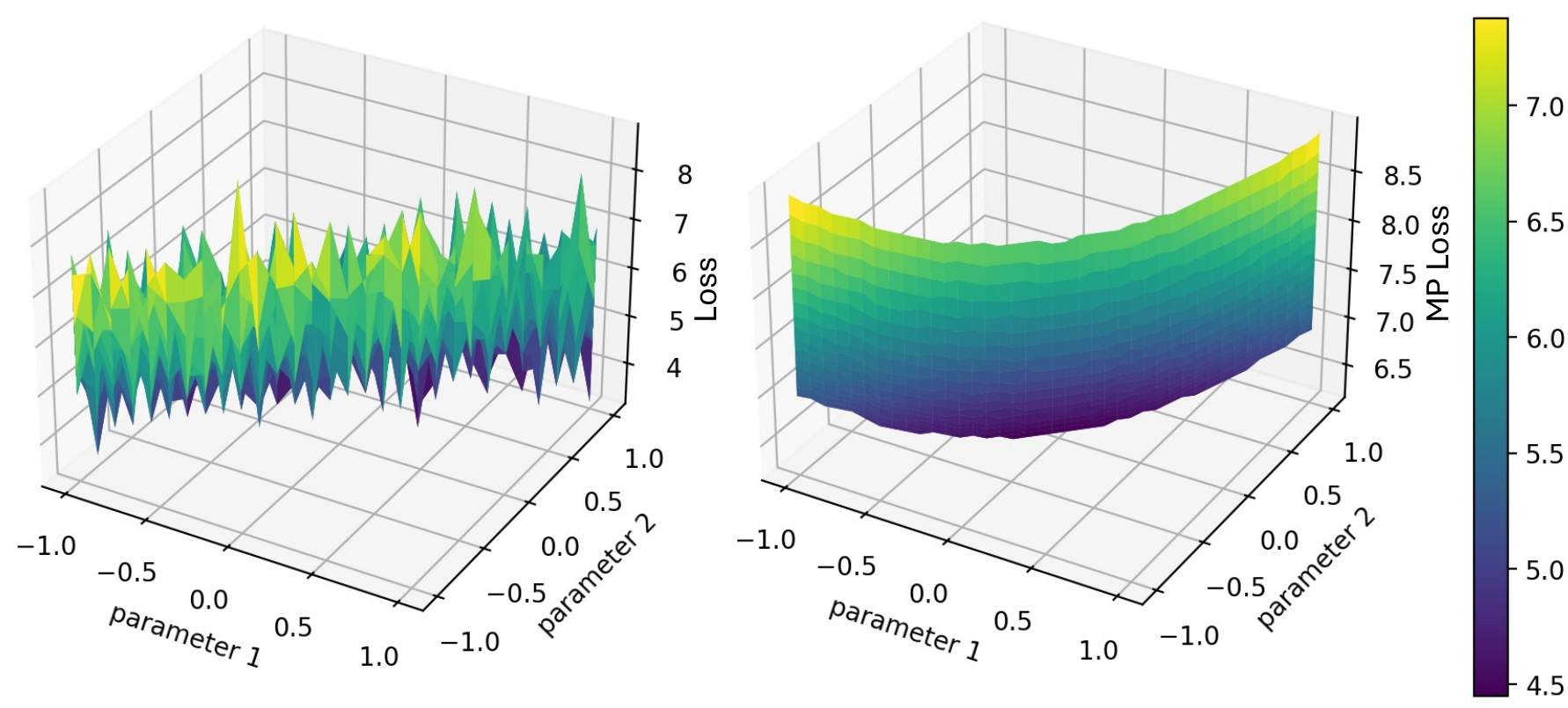}
    \caption{Loss landscape comparisons for the vanilla and MP objective functions for a controlled Lorenz system given by Equation \ref{lor_cont_2}. {The values of the loss functions are not important here as we highlight that the MP loss landscape is smooth and convex compared to the standard loss.}}
    \label{fig:loss land}
\end{figure}

To illustrate how the MP optimization improves optimization further, we explore the loss landscape of both vanilla and MP objective functions for a benchmark problem introduced in Chung et al.(2022)~\cite{chung2022optimization}.
\begin{equation}
    \begin{aligned}\label{lor_cont_2}
        &\frac{dx}{dt}=\sigma(y-x)\\
        &\frac{dy}{dt}=x(\rho-z)-y\\
        &\frac{dz}{dt}=xy-\beta z + f(t),
    \end{aligned}
\end{equation}
The system is integrated from {$t_i=0$ to $t_f=20$} with 2000 time-steps. {$f(t):(t_i,t_f)\longrightarrow \mathbb{R}$ is a time varying controller aimed at minimizing the objective function. It is taken as a vector of 2000 parameters defined at each time-step.}  The objective function for optimization is defined as 
\begin{equation}\begin{aligned}\mathcal{J}&=\frac1{t_f-t_i}\int_{t_i}^{t_f}\mathcal{I}[\mathbf{q}] dt,\\\mathcal{I}[\mathbf{q}]&=\begin{cases} \frac12\left(\frac{2x+y}5\right)^2&2x+y\geq0\\ 0&\text{otherwise.}\end{cases}\end{aligned}\end{equation}
Chung et al.(2022) showed that MP optimization can reduce the objective to 99.9\% as compared to vanilla optimization which can reduce it to only 44.6\%~\cite{chung2022optimization}. Figure \ref{fig:loss land} shows the comparison of loss landscape for vanilla and MP objective function varying two arbitrary parameters of the time-varying control $f(t)$. It clearly shows that the vanilla loss function is highly non-convex which hinders convergence during optimization.

In past work, Chung et al.(2022) have use MP optimization to optimize 2D and 3D chaotic turbulent flows~\cite{chung2022optimization}. In our experiments for this article, we extend it to learning of chaotic systems using NODE. Computational resources used for all the experiments was one 40-GB NVIDIA A100 GPU.

\subsection{Kuramoto Sivashinsky equation}\label{KS_problem}

The Kuramoto Sivashinsky(KS) equation is given by a fourth-order non-linear partial differential equation,
\begin{equation}\begin{aligned}
\frac{\partial q}{\partial t}=-q\frac{\partial q}{\partial x}-\frac{\partial^2q}{\partial x^2}-\frac{\partial^4q}{\partial x^4}.
\end{aligned}\end{equation}
Solutions to the KS equation show chaotic dynamics which eventually converge onto a low-dimensional inertial manifold over long periods~\cite{foias1986inertial}. Predicting the long term trajectories of such chaotic systems is very challenging as small errors have large effects on the roll-outs after multiple Lyapunov time lengths. So, we aim to have a model that can predict the KS dataset accurately in short times while preserving invariant statistics for long term predictions. Our training dataset comprises a single trajectory that has converged onto the inertial manifold, covering the time span from t = 0 to t = $10^5$, sampled every 0.25 time units. A single long trajectory can be split into multiple smaller trajectories of different initial conditions due to the ergodic behavior of KS system~\cite{hyman1986order}. The training conditions are taken identical to Stabilized NODE paper~\cite{linot2023stabilized}.
\par
\begin{figure}[hp]
    \centering
    \includegraphics[width=0.99\textwidth]{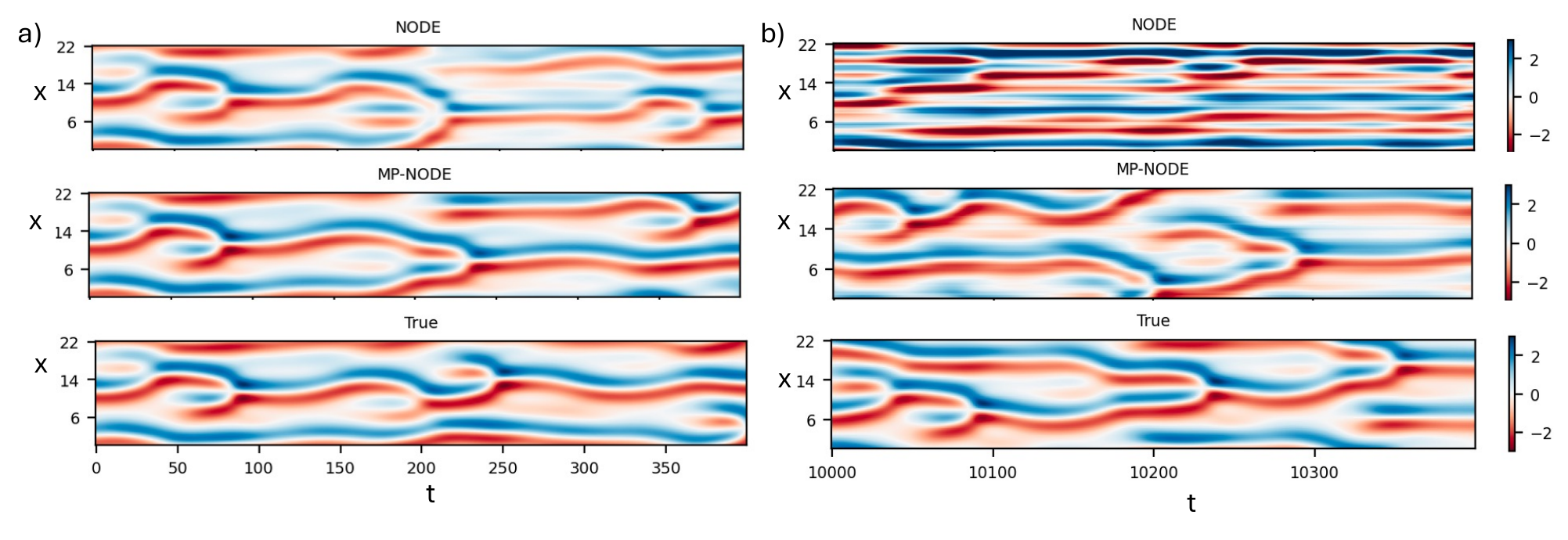}
    \caption{A comparison for NODE vs MP-NODE for KS equation for (a) short term (b)long term}
    \label{fig:ks_time}
\end{figure}
\begin{figure}[hp]
    \centering
    \includegraphics[width=0.7\textwidth]{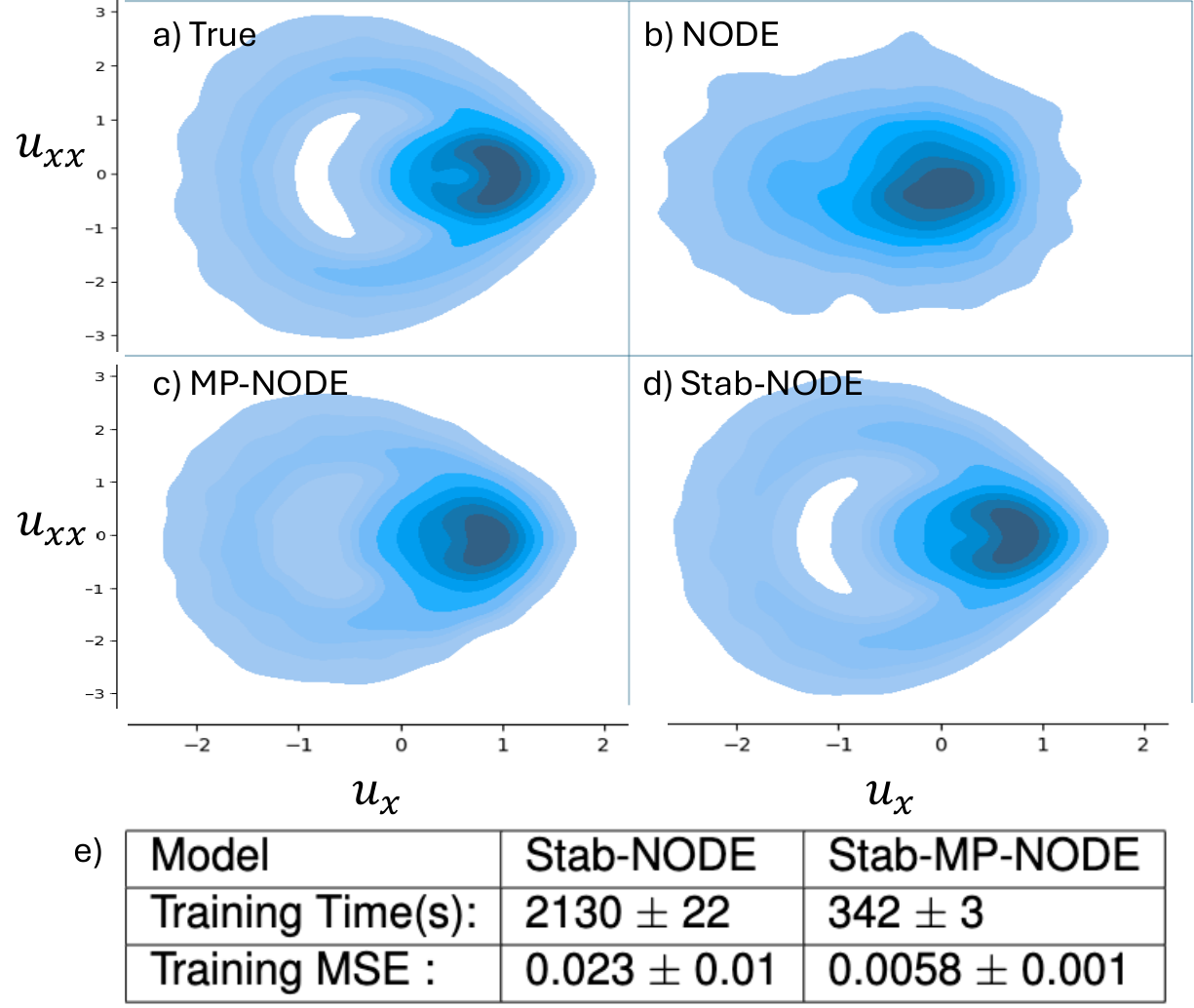}
    \caption{Comparison of joint PDF between first and second derivative of the state vectors for (a) Ground Truth, (b) NODE (c) MP-NODE and (d) Stabilized NODE predictions (e) Training time comparison for Stabilized MP-NODE and Stabilized NODE of a single batch for optimization.}
    \label{fig:ks_pdf}
\end{figure}
Figure~\ref{fig:ks_time} shows the time evolution of the vanilla NODE and MP-NODE models compared to the ground truth.
It is observed that, for test data, in the short term (t $\approx$ 50) both NODE and MP-NODE can accurately predict the ground-truth, but deviations emerge soon after due to the chaotic nature of the system. Such an observation is typical for the KS system which possesses a Lyapunov time-scale of $ \tau_L \approx $ 22.  For longer predictions, the MP-NODE maintains the features of the inertial manifold to over 30 Lyapunov times, whereas the NODE prediction significantly deviates from it.
We qualitatively access the inertial manifold here with the joint probability distribution function (PDF) of the first and second derivative of the state vectors following related works\cite{linot2023stabilized}. We further compare the Kullback Leibler(KL) divergence of the joint PDFs for different models with varying hyper-parameters in Appendix \ref{App ablation} for quantitative assessment. Figure~\ref{fig:ks_pdf} shows the mentioned joint PDFs after the evolving state has reached the attractor. The PDF is taken up to $t=750$ (beyond which the standard NODE diverges).
It is also compared with the stabilized NODE architecture \cite{linot2023stabilized} which explicitly introduces a linear term in the RHS of Equation \ref{NODE} to make the prediction stable. The PDF from MP-NODE agrees well with the ground-truth PDF. Although we observe that the stabilized NODE predicts the statistics closest to the ground truth, we note that the results from the MP-NODE technique have been obtained with \emph{no underlying assumptions} about the structure of the dynamical system. Moreover, the MP technique is a general algorithm for learning time-series data, and it can be incorporated into the stabilized NODE as well for faster convergence and improved stability.

\subsubsection{Return Period}
\begin{figure}[hp]
    \centering
    \includegraphics[width=0.7\linewidth]{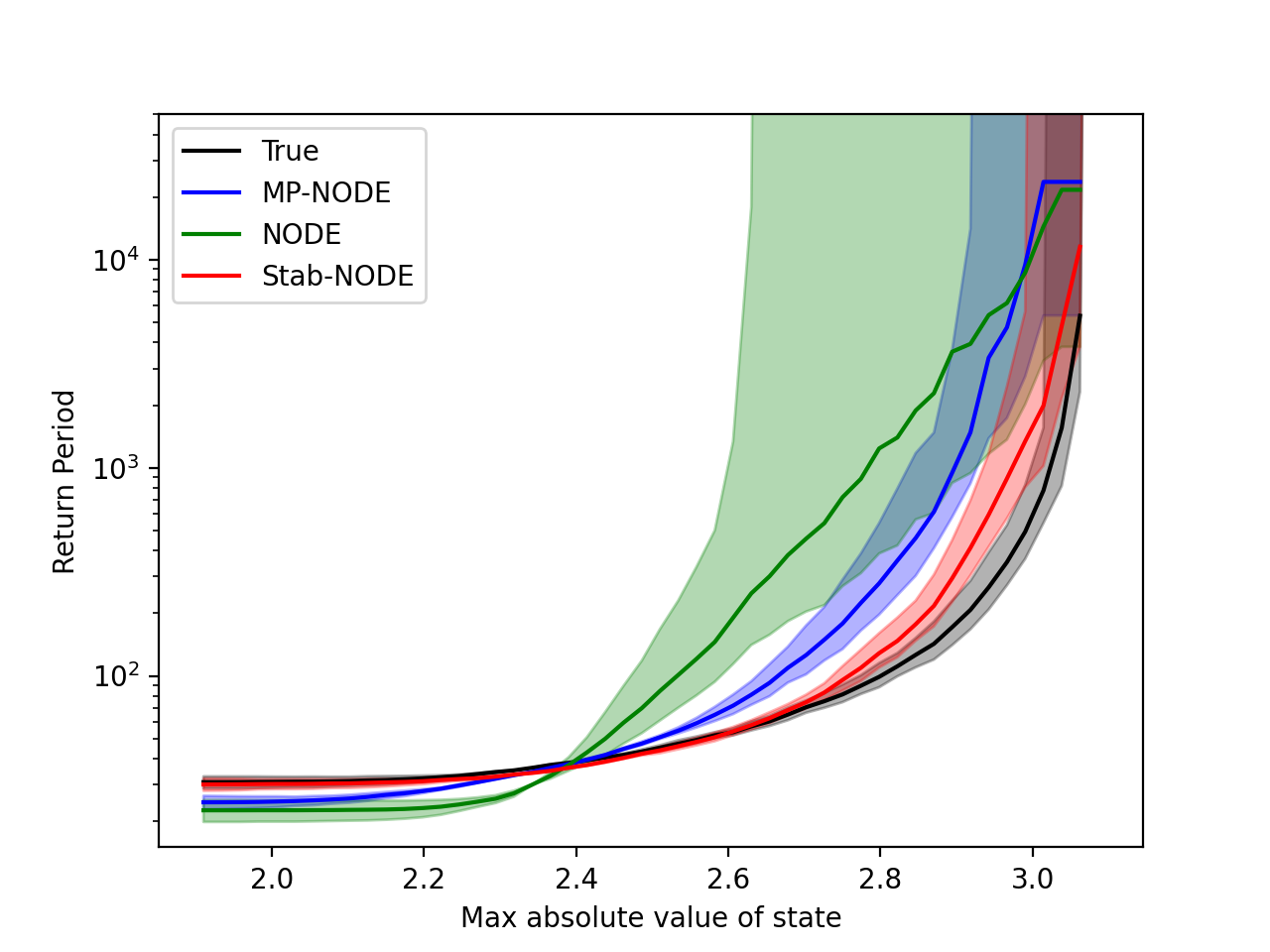}
    \caption{Return period comparisons between NODE,MP-NODE and stab-NODE with ground truth. The vertical axis represents the expected time-period for the dynamical system to return to a maximum value of the state given by the horizontal axis.}
    \label{fig:return}
\end{figure}

{A return period is a statistical measure that estimates the average interval of time between the recurrence of events corresponding to specific magnitudes. It is computed as
\begin{align}
    T=\frac1N\sum_{i=1}^{N-1}(t_{i+1}-t_i)
\end{align}
where $t_i$ corresponds to one occurrence of an event and $t_{i+1}$ refers to its next. In a return period plot, we compute the average intervals between events (plotted on the vertical axis) for the maximum value of the state to be higher than a specific magnitude (plotted on the horizontal axis). Specifically, we used 3000 rollouts of a test initial condition to calculate the return period. Figure \ref{fig:return} shows that the MP-NODE matches the return period of the training data more closely compared to vanilla NODE. The stab-NODE matches the ground truth better as it is physics-based. However, this shows the ability of MP-NODE to recover a dynamical system that generates extreme events (here denoted by large magnitudes of the state) at more accurate intervals than a regular NODE.  }

\subsection{Kolmogorov Flow}\label{subsec:kolmogorov}
In this section, we study the performance of the proposed framework for two-dimensional homogeneous isotropic turbulence with Kolmogorov forcing governed by the incompressible Navier-Stokes equations. The purpose of these experiments is to assess the capabilities of MP-NODE for a significantly high-dimensional state as well as a large number of parameters. Forced two-dimensional turbulence exhibits the classical characteristics of chaotic dynamics and has become a popular benchmark for ML techniques used for the prediction of dynamical systems~\cite{stachenfeld2021learned,schiff2024dyslim,shankar2023differentiable}. 
The 2D Navier-Stokes Equation is given by 
\begin{equation}\begin{aligned}\frac{\partial\mathbf{u}}{\partial t}+\nabla\cdot(\mathbf{u}\otimes\mathbf{u})&=\frac{1}{Re}\nabla^2\mathbf{u}-\frac{1}{\rho}\nabla p+\mathbf{f}\\\nabla\cdot\mathbf{u}&=0,\end{aligned}\end{equation}
where $\mathbf{u}=(u,v)$ is the velocity vector, $p$ is the pressure, $\rho$ is the density, $Re$ is the Reynolds number, and f represents a forcing function given by 
\begin{equation}
\mathbf{f}=A\mathrm{sin}(ky)\hat{\mathbf{e}}-r\mathbf{u}
\end{equation}
parameterized by the amplitude $A=1$, wavenumber $k=4$, linear drag $r=0.1$, and the Reynolds number $Re=1000$ chosen for this study~\cite{shankar2023differentiable}. $\hat{\mathbf{e}}$ denotes the x-direction unit vector. The initial condition is taken as a random divergence free velocity field~\cite{Kochkov2021-ML-CFD}.
The ground truth datasets are obtained from direct numerical simulation(DNS)~\cite{kochkov2021machine} of the governing equations in a doubly periodic square domain with L = 2$\pi$ discretized in a uniform grid (512x512) and filtered to a coarser grid (64x64). Details for constructing this dataset can be found in the work by Shankar et al.(2023) \cite{shankar2023differentiable}. The trajectories are temporally sampled after flow reaches the chaotic regime with a gap of $T = 256\Delta t_{DNS}$
to ensure sufficient distinction between two snapshots empirically.
\begin{figure}[!h]
    \centering
    \includegraphics[width=\textwidth]{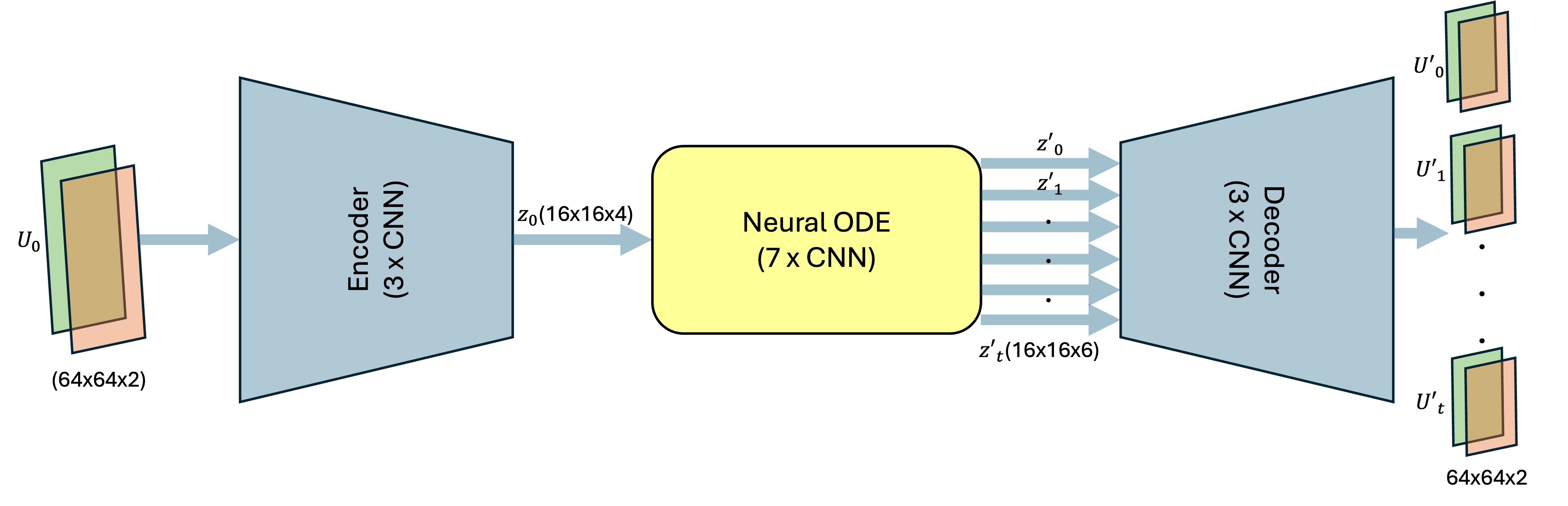}
    \caption{Encoder-NODE-Decoder Architecture for prediction of two-dimensional homogeneous isotropic turbulence with Kolmogorov forcing.}
    \label{fig:AE_node}
\end{figure}

For the ML framework to predict these dynamics, we utilize an Encoder-NODE-Decoder architecture, as shown in figure \ref{fig:AE_node}. The input is an initial condition with 2 channels ($u,v$) of a 64x64 grid. The encoder layer has 3 Convolutional Neural Network (CNN) layers with varying kernel sizes (7,5,2) and number of channels (8,16,4). {The latent initial condition ($z_0$ in Figure \ref{fig:AE_node})} has 4 channels which are augmented~\cite{dupont2019augmented} with 2 extra channels to allow the NODE to learn complex dynamics in a low-dimensional latent space. The NODE has 7 CNN layers with kernel size 3 and 16 number of channels (except the output layer). It also has dilated kernels~\cite{stachenfeld2021learned} with varying dilation (1,2,3,4,3,2,1) to capture long range interactions. The dilated CNN is similar to standard CNN but has kernels with holes that can skip pixels, thereby increasing the receptive field without amplifying the number of parameters. The latent-space NODE outputs are rolled out autoregressively for $t$ time-steps given a latent initial condition. Each NODE output timestep is passed to the decoder to reconstruct a physical space flow-field. {The integrator and activation function used are Tsit5~\cite{tsitouras2011runge} and 'gelu' respectively.}
\begin{figure}[!h]
    \centering
    \includegraphics[width=0.95\textwidth]{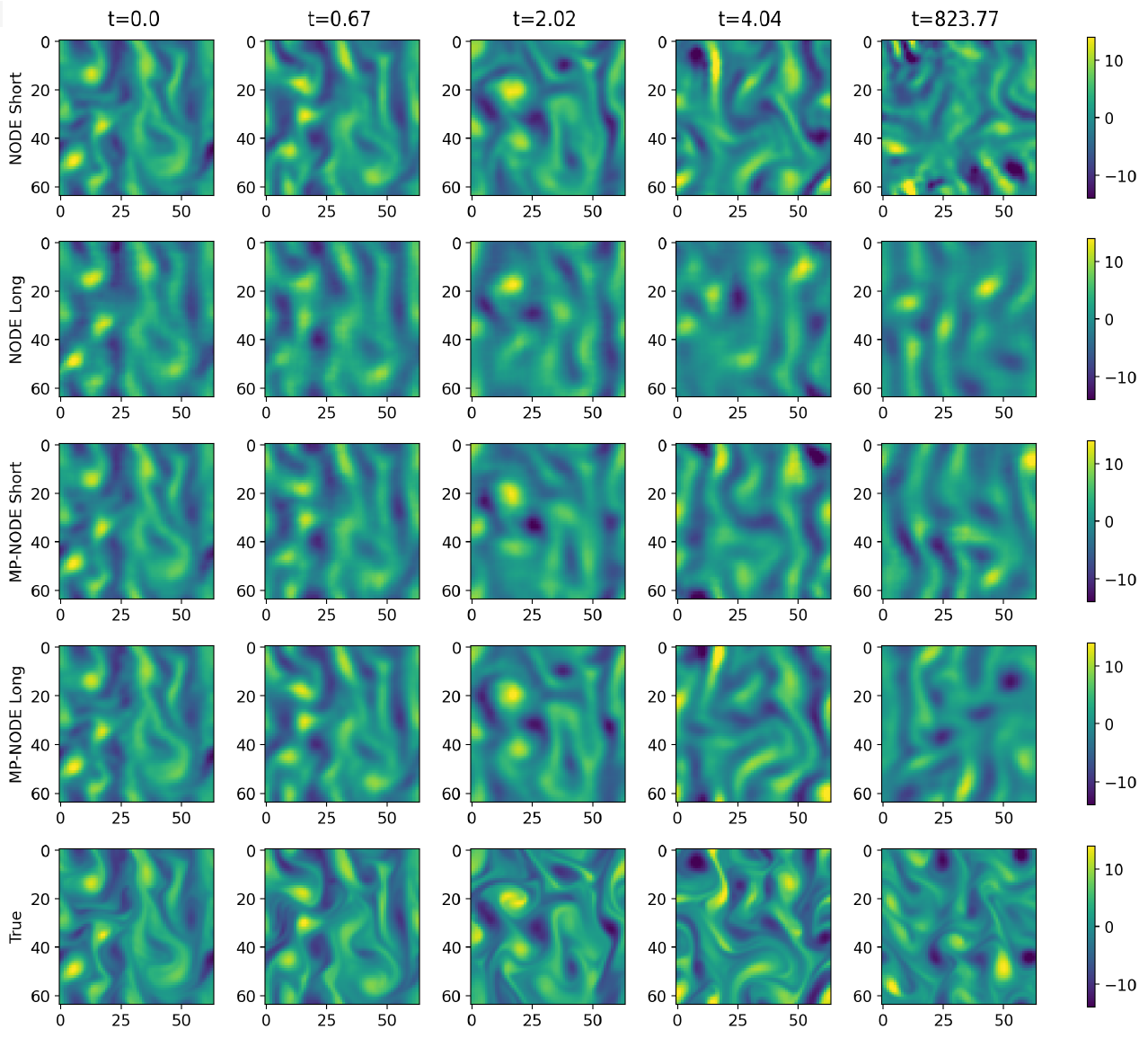}
    \caption{Vorticity predictions for the different NODE algorithms through autoregressive rollouts. Here `Short' and `Long' signify backpropagating through short and long rollouts during training respectively.}
    \label{fig:KF_vort}
\end{figure}

The proposed model is trained with the Adam optimizer with cosine scheduling and learning rates starting at $10^{-3}$ and reducing to $10^{-5}$. We explore two lengths of rollouts - short and long with 5 and 61 time-steps respectively. For the MP-NODE the longer trajectory is split into 12 equal domains penalizing the local discontinuity with penalty strength varying from $10^{-5}$ to $100$. Ideally, the discontinuity points should be based on the Lyapunov time scale of the system keeping the integration time length smaller than the Lyapunov time. However, the Lyapunov time for a NODE changes {throughout the} optimization, as it is learning the characteristics of the target chaotic system. So, we choose the discontinuity points empirically using fewer points to start with and gradually increase them if slow convergence is observed. All models are trained for the same duration to compare performance fairly. The loss function used for training is the mean square error between the predicted and actual flow fields. 

Furthermore, we quantify the uncertainty of our ML predictions for this particular experiment using the Gaussian stochastic weight averaging technique \cite{morimoto2022assessments,izmailov2018averaging}. Specifically, the model is trained up to convergence, the optimizer is switched to stochastic gradient descent (SGD) with a constant learning rate, and each set of neural network weights obtained after 10 epochs of SGD is saved as a different model. Predictions from these 10 different models are used for ensemble statistics.
\begin{figure}[hp]
    \centering
    \includegraphics[width=\textwidth]{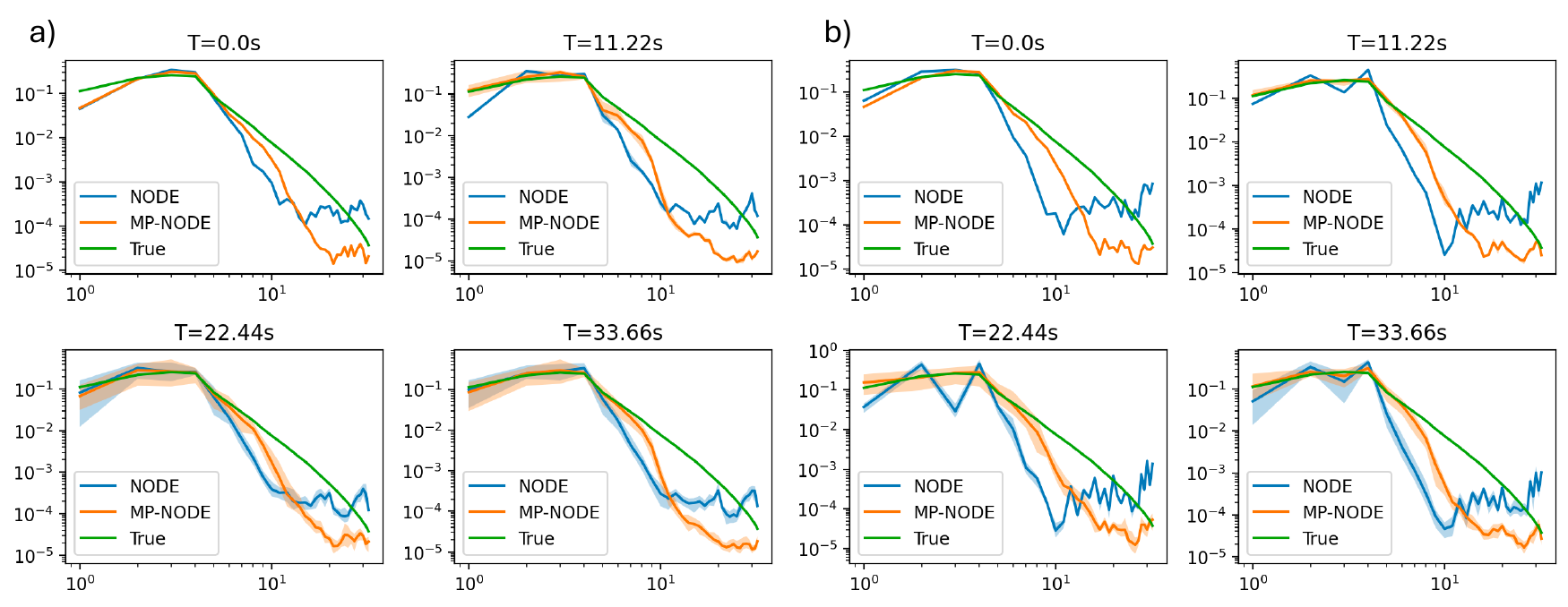}
    \caption{Energy spectrum comparison between NODE and MP-NODE trained for a) short rollouts b) long rollouts}
    \label{fig:KFspec}
\end{figure}
\begin{figure}[hp]
    \centering
    \includegraphics[width=0.75\textwidth]{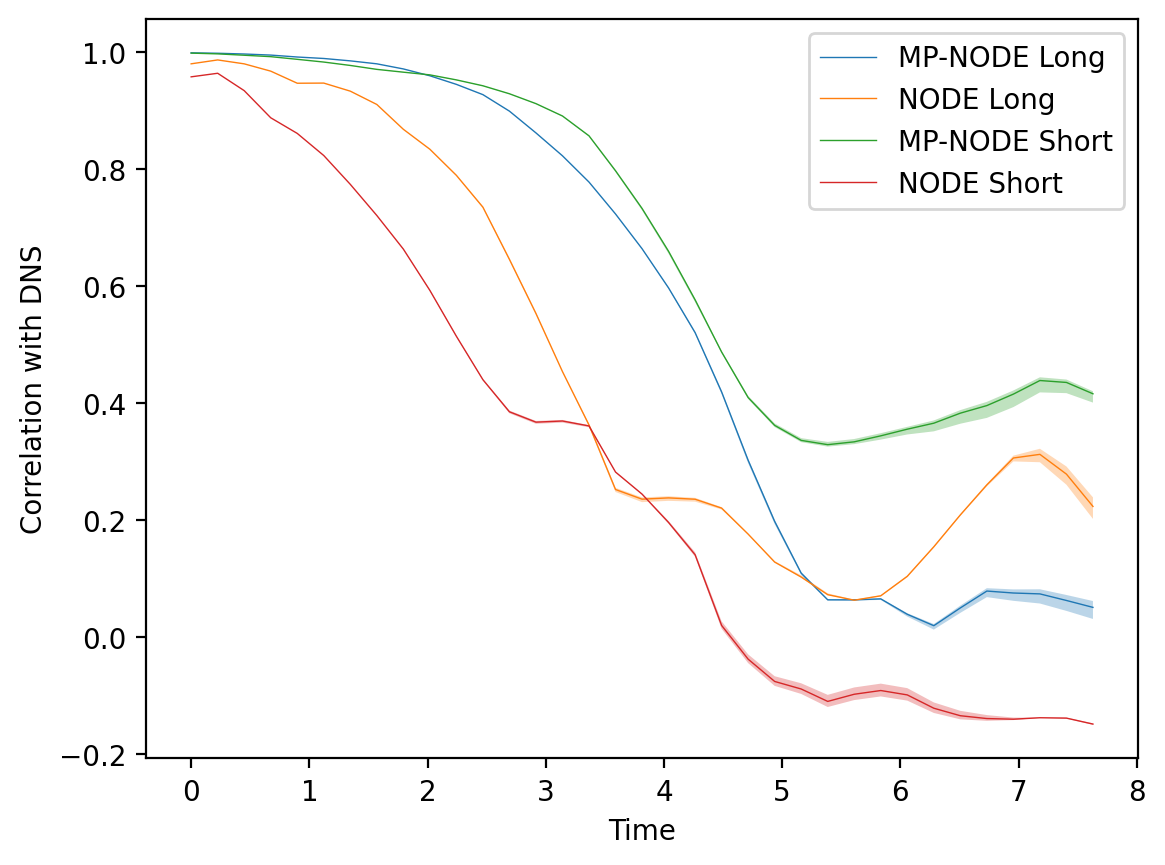}
    \caption{Correlation with DNS flow field for training with short and long rollouts for NODE and MP-NODE}
    \label{fig:corr_KF}
\end{figure}
\begin{figure}
    \centering
    \includegraphics[width=0.75\textwidth]{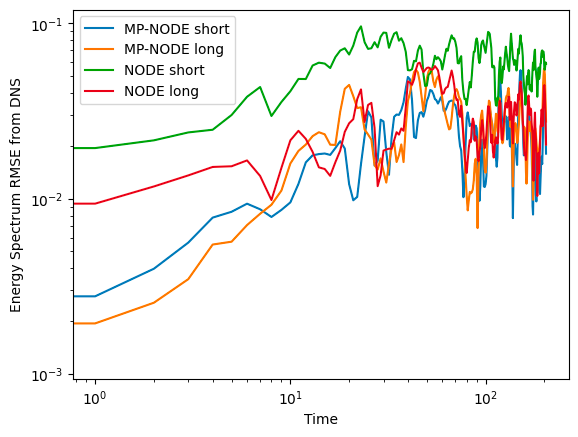}
    \caption{Error comparisons using Equation \ref{rmse} between true and predicted energy spectrum of the flow fields for various models.}
    \label{fig:rmse_KF}
\end{figure}

We first assess the predictions by comparing the angle-averaged kinetic energy spectrum in wavenumber space. Figure \ref{fig:KFspec} shows that the MP-NODE has a better overlap with the true energy spectrum for both short and long rollouts, particularly in the inertial range. It is also observed that the variance in predictions for the kinetic-energy spectra grows with increasing time, particularly at the lower wavenumbers. Regular NODE predictions show a significant deviation from the true-scaling of the spectra in the form of pile-up error near the grid cut-off. This is improved by the use of the MP-NODE method. Secondly, Figure \ref{fig:corr_KF} shows that MP-NODE performs significantly better than vanilla NODE. It has a high correlation with DNS for much longer than vanilla NODE. For MP-NODE, training with short rollouts performs slightly better as correlation is a short term metric for chaotic systems. Furthermore, NODE training with long rollouts performs better than short rollouts. This analysis demonstrates the improvement imparted by using the MP optimization process.
We use a mean-squared error metric to quantify the deviation from ground truth energy spectrum at each timestep of the test rollout. It is computed as
\begin{equation}\label{rmse}
    \sqrt{\frac{1}{K}\sum_{k=1}^K\|E(\hat{k}) - E(k)\|^2}
\end{equation}
where $E(\hat{k})$ is the predicted energy spectrum and $E({k})$ is the energy spectrum of the DNS flow field. Figure \ref{fig:rmse_KF} shows that training with long rollouts preserve the energy spectrum better for both models.
We can conclude that the use of the MP-NODE improves the stable prediction of chaotic dynamical systems while preserving their invariant statistics. 

\subsection{Learning a climate emulator using the ERA 5 dataset}\label{subsec:era5}
\begin{figure}[!h]
    \centering
    \includegraphics[width=0.79\linewidth]{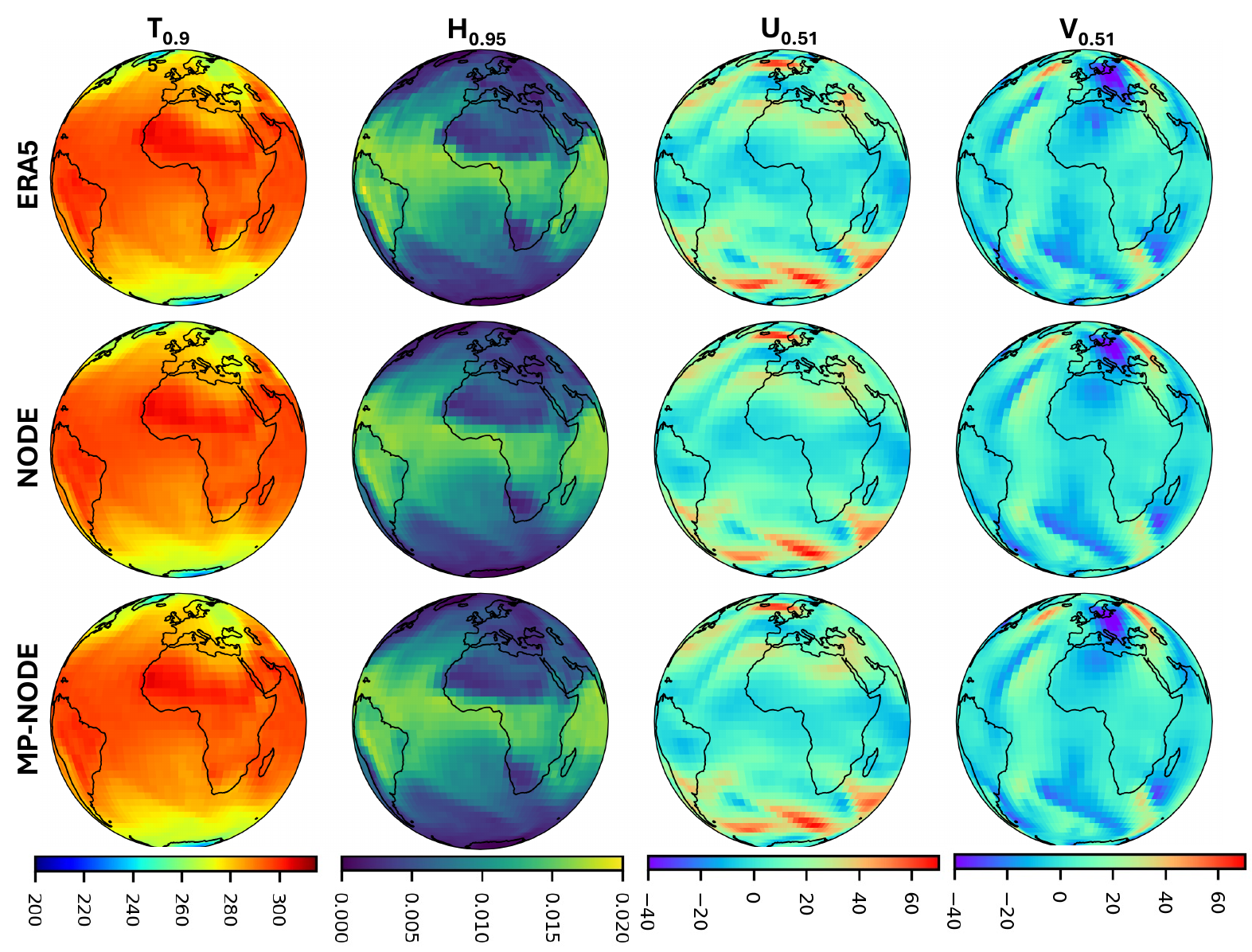}
    \caption{Day 1 prediction of global weather by ERA5, NODE and MP-NODE respectively (from top to bottom).}
    \label{fig:Day 1 comparison ERA5}
\end{figure}
For applications to a more complex, multivariate dataset, we train our model on ERA5 \cite{Hersbach2020} from January 1, 2000, to December 15 2009. The ERA5 dataset is an observation-based reanalysis of the atmosphere, which we have regridded to a T30 Gaussian grid and vertically interpolated onto $\sigma$-levels (see \cite{Arcomano_2022} Section 2.5 for details on the dataset). This grid corresponds to the same horizontal, vertical resolution, and $\sigma$-levels as the LUCIE model \cite{guan2024lucie}. We use temperature (t) and specific humidity (q) at $\sigma_{0.95}$ and zonal(u) and meridional(v) wind at $\sigma_{0.51}$. We incorporate total incoming solar radiation (TISR) as a temporal embedding within the model and is also a boundary condition used by numerical-based climate models. With the intention of studying both the short-term forecasting ability and long-term behavior and stability of each model, we compare the predictive capabilities of NODE and MP-NODE to produce accurate 14-day forecasts as well as realistic, long term climate during multi-year simulations. We note stable, ML-based climate emulators (e.g., \cite{watt2023ace}) have only been demonstrated recently and can prove to be quite challenging from a machine learning-perspective. 
\begin{figure}[!h]
    \centering
    \includegraphics[width=0.78\linewidth]{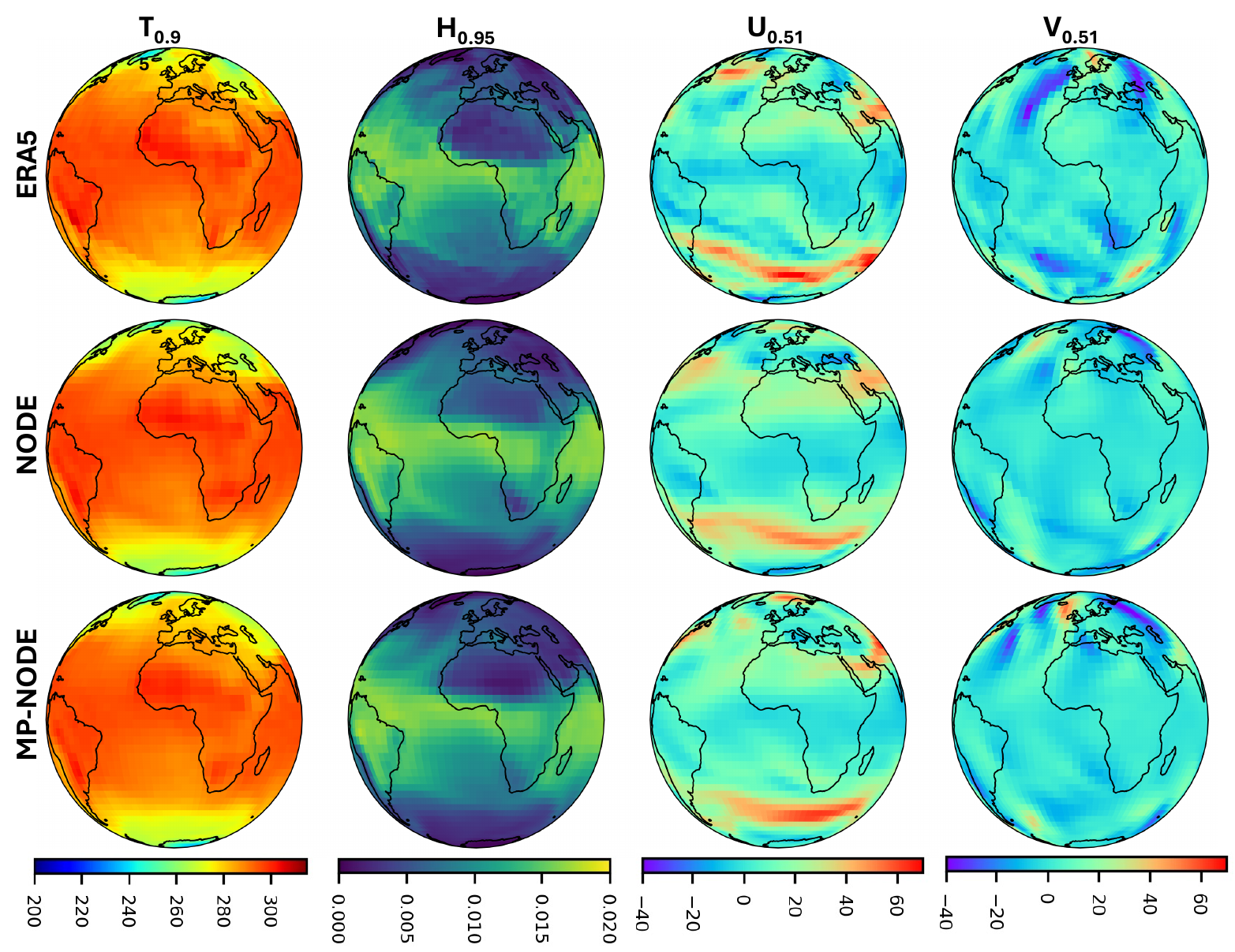}
    \caption{Day 14 prediction of global weather by ERA5, NODE and MP-NODE respectively  (from top to bottom).}
    \label{fig:Day 14 comparison ERA5}
\end{figure}
\begin{figure}[!h]
    \centering
    \includegraphics[width=0.7\linewidth]{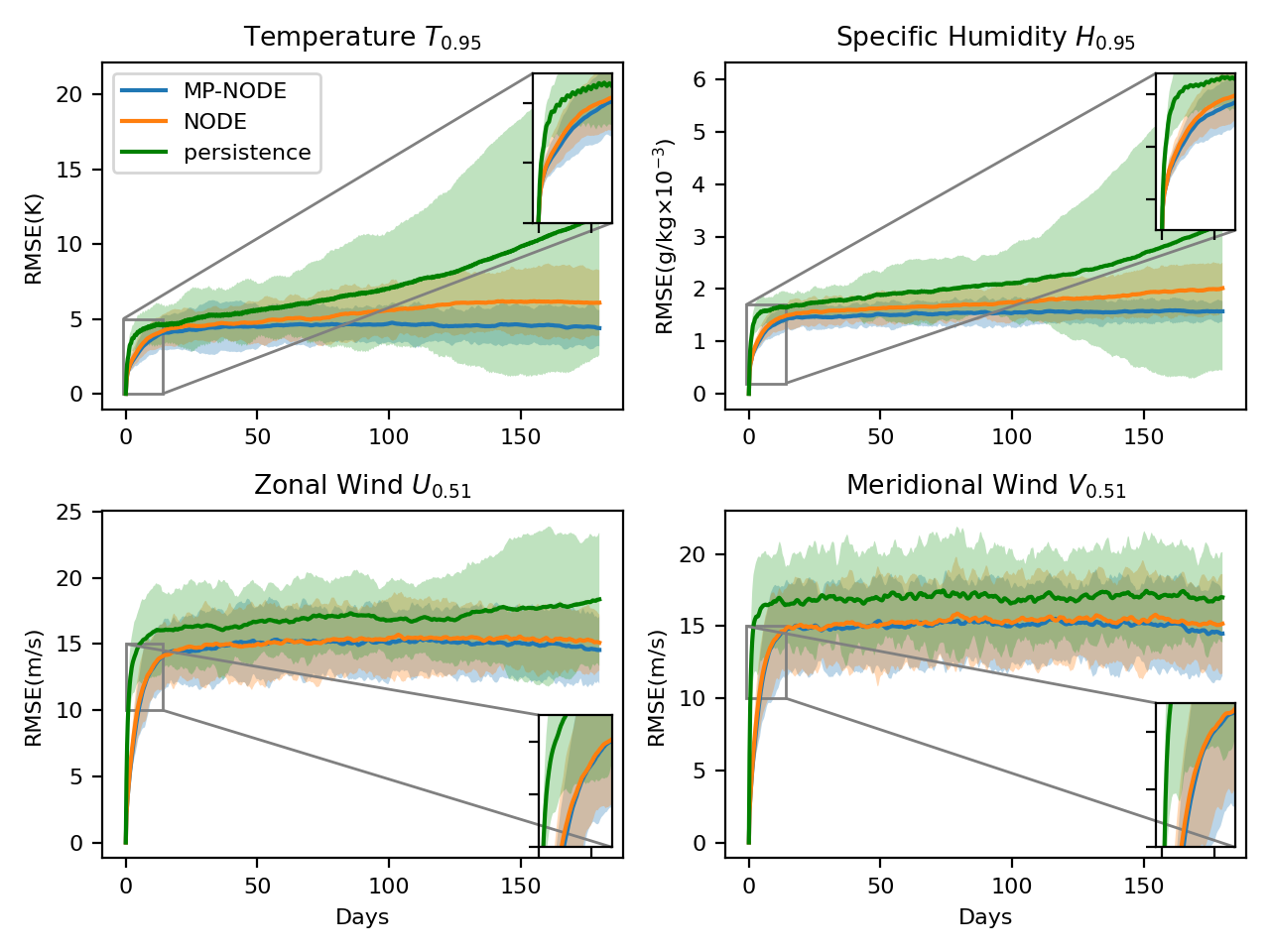}
    \caption{Root Mean Square Error(RMSE) from ground truth comparison for persistence, NODE and MP-NODE. The ensemble is performed using random 50 initial conditions between 2011 to 2016 and the shaded region represents $\pm \ 3\ \times$ standard deviation.}
    \label{fig:RMSE ERA5}
\end{figure}
\begin{figure}[!h]
    \centering
    \includegraphics[width=0.7\linewidth]{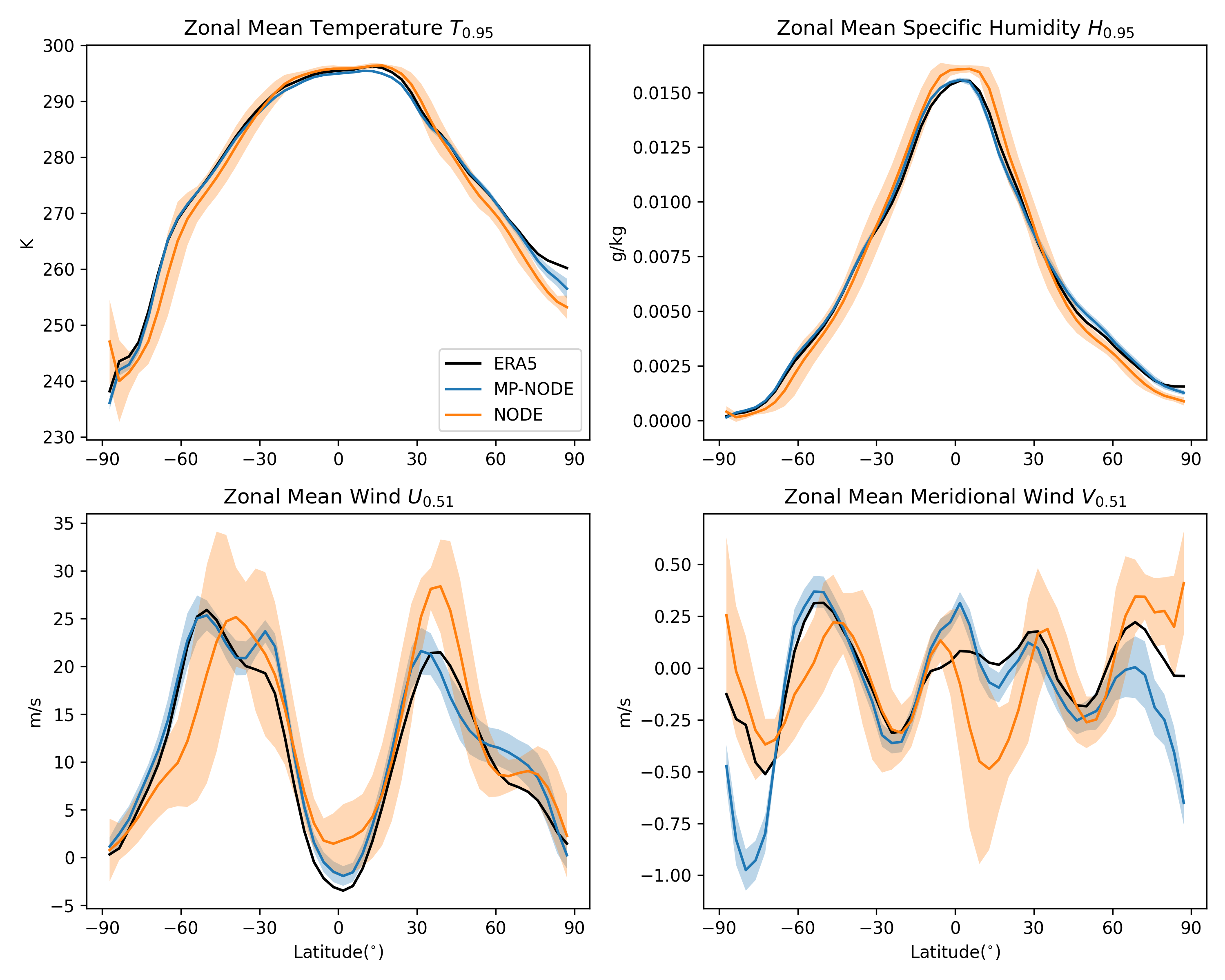}
    \caption{1 year time and zonal mean climatology comparison for NODE and MP-NODE. The ensemble is performed using random 50 initial conditions between 2011 to 2016 and the shaded region represents $\pm \ 3\ \times$ standard deviation.}
    \label{fig:climatology ERA5}
\end{figure}
The architecture used is similar to Figure \ref{fig:AE_node} with the exception of absence of a compressive encoder to mitigate loss of information. We have used techniques like push-forward trick\cite{brandstetter2022message} and Euler integration to enhance memory efficiency.  
\begin{figure}[h]
    \centering
    \includegraphics[width=0.7\linewidth]{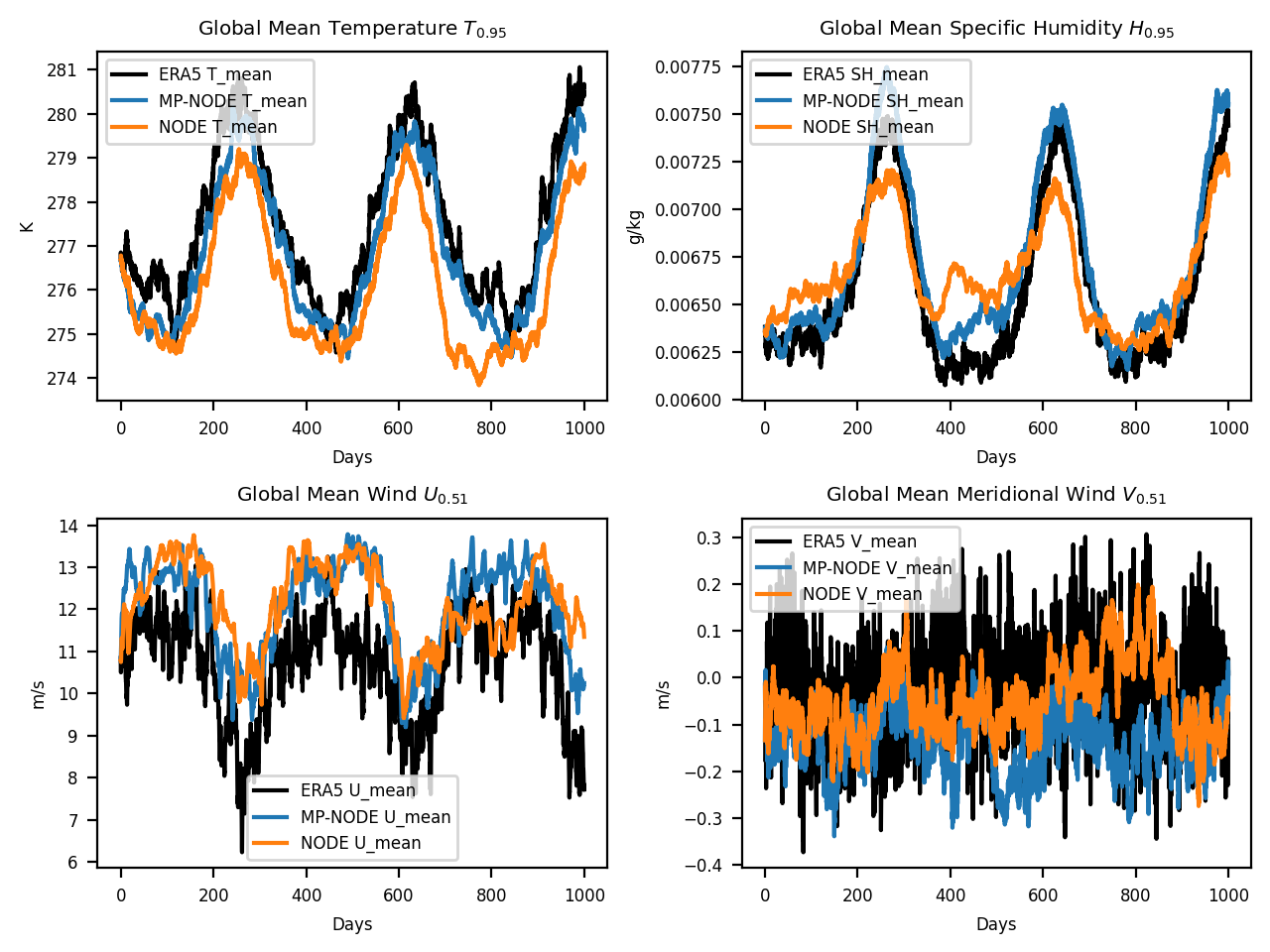}
    \caption{Comparison of global Mean of prognostic variables for 1000 days for NODE and MP-NODE.}
    \label{fig:seasonal variation}
\end{figure}
The short-term prediction starts from 50 different initial conditions in the testing period and is subjected to various climatology tests to examine the advantages of MP optimization. The reconstructions of global weather variables for a particular Day 1 (Figure \ref{fig:Day 1 comparison ERA5}) and Day 14 (Figure \ref{fig:Day 14 comparison ERA5}) are very close to the ground truth. By Day 7, MP-NODE starts to show its advantages over vanilla NODE, especially for zonal and meridional winds. We also compare the Root Mean Square Error (RMSE) from ERA5 against NODE, MP-NODE, and persistence. The persistence model is considered an elementary forecasting model to compare the accuracy of other models. It assumes that the weather is static and does not change with time. Figure\ref{fig:RMSE ERA5} shows that MP-NODE beats both persistence and vanilla NODE for temperature and specific humidity. For zonal and meridional wind, it beats the persistence model and has approximately similar RMSE as the vanilla NODE. However, it is observed that vanilla NODE performs worse than persistence over longer periods of time. 

Secondly, we compare the long-term climatology using time and zonal mean of the prognostic variables in Figure \ref{fig:climatology ERA5}. We observe that the MP-NODE matches the zonal mean of ERA5 variables across latitudes better than vanilla NODE. MP-NODE is able to correctly capture the jet structure of the zonal wind while the vanilla NODE produces an overactive jet that is too close to the equator in each hemisphere. The vanilla NODE also has a much too warm climatology over the Antarctica region, where MP-NODE closely matches ERA5. While the deviation is larger for meridional wind especially towards the poles, this may at least partially be explained by the fact that we do not use any inductive biases during training to incorporate the spherical geometry of the earth. 

Finally, we compare the global mean of prognostic variables for 1000 days to observe the seasonal variations in Figure \ref{fig:seasonal variation}. The MP-NODE both matches the global means of ERA5 much closely and produces more temporal variability than vanilla NODE, further showcasing its advantages in long-term predictions. This model still uses a lightweight version of ERA5 data and the implementation to high resolution ERA5 for all variables is left for future research.

\section{Discussion}\label{d and c}

\subsection{Limitation and Future Works}

In this section, we discuss some limitations of the proposed methodology. {We identify the increased memory requirement of MP-NODE due to the increase in discontinuity points(trainable parameters) for the longer rollouts.} One approach to address this issue is to learn dynamics on a low-dimensional manifold - however, this may cause the loss of small scale information on the flow field. This is also observed in our Kolmogorov flow experiments as evident in Figures \ref{fig:KF_vort} and \ref{fig:KFspec}, where finer structures are smoothed. For systems where this is undesirable, future work can investigate memory-efficient adjoint sensitivity methods \cite{djeddi2019fdot,djeddi2020memory,zhang2022memory,zhuang2021mali}. {This is not a very significant limitation because the total memory requirement is of the same order as vanilla NODE}. A second limitation is the computation of stochastic gradients within the MP-NODE framework. The proposed method specifies the magnitudes of discontinuities as learnable parameters for optimization. Consequently, every time a new batch is taken for gradient computation, these parameters have to be reinitialized. This causes slow and sub-optimal convergence for the discontinuity loss. We observed this effect more significantly for the Kolmogorov-flow {and ERA5} case where the state-space dimensionality (directly related to the dimension of the number of discontinuity parameters) is high. An intuitive solution to this can be to keep a global register for intermediate initial conditions and initialize the batches based on updated values from the register. However, this will require an extremely high number of learnable parameters in the register for large datasets which will complicate the scalability of the proposed algorithm. 

\subsection{Conclusion}

The advent of neural differential equations has opened new avenues by combining the flexibility of neural networks with the accuracy of numerical solvers, allowing for improved predictability of dynamical systems. However, the challenge of long-term stable prediction for chaotic systems has mostly remained unsolved.
Learning chaotic dynamics is challenging as the optimization problem becomes extremely non-convex, making the standard gradient useless.
Existing methods like least-squares shadowing, while attractive, suffer from very high computational costs for high-dimensional systems as well as overparameterized models. The algorithm proposed in this work solves this issue by dividing the time domain into multiple steps and allowing a penalized discontinuity to calculate sensible gradients with linear computational complexity. The benefit of MP-NODE is that it can be extended to several time-series forecasting techniques that rely on multistep loss calculations and backpropagation through time. Moreover, this advancement has the potential to improve the understanding and predictability of chaotic systems and has broader implications for several scientific domains.

\section*{Acknowledgments}
This material is based upon work supported by the U.S. Department of Energy (DOE), Office of Science, Office of Advanced Scientific Computing Research, under Contract No. DE-AC02–06CH11357. 
RM and DC acknowledge support from DOE ASCR award ``Inertial neural surrogates for stable dynamical prediction" and DOE FES award ``DeepFusion Accelerator for Fusion Energy Sciences in Disruption Mitigation". RM and DC also acknowledge the support of Penn State ICDS computing resources. This work was performed in part under the auspices of the U.S. Department of Energy by Lawrence Livermore National Laboratory under Contract DE-AC52-07NA27344. LLNL-JRNL-862644. {We also acknowledge Haiwen Guan from Pennsylvania State University for his assistance with the ERA5 benchmark experiments.}

\bibliographystyle{unsrt}  
\bibliography{MP_NODE.bib}  

\section{Appendix}
\subsection{Algorithms}\label{Algo_Sec}
\begin{algorithm}[h!]\label{Algo2}
\caption{Multi-step NeuralODE - Formulation 2}\label{alg:cap}
\begin{algorithmic}
\For{$\mu$ in  \{$\mu_i$\}} 
    \While{ $j<$  n\_batches}
    \State get\_batch($t, \textbf{q}$, ..) \Comment{Get a new batch}
    \State Initailize $\boldsymbol{q_k}$ \Comment{Initialize discontinuities for the new batch}
    \State Params $\gets$($\boldsymbol{\Theta}$, $\boldsymbol{q_k}$) \Comment{Append neural network parameters and $\boldsymbol{q_k}$ as set of trainable parameters}
    \While{$k<$ num\_iters}
    \State ($\nabla \mathcal{L}_{\boldsymbol{\Theta}}, \nabla \mathcal{L}_{\boldsymbol{q_k}}$) = compute\_gradients($\mathcal{L}$,Params,$\mu_i$) \Comment{Compute gradients using Adjoint or autograd}
    \State Params $\gets$ optimizer.step(Params,$\nabla \mathcal{L}_{\boldsymbol{\Theta}}, \nabla \mathcal{L}_{\boldsymbol{q_k}}$) \Comment{Update the trainable parameters}
    \EndWhile
    \EndWhile
\State update $\mu$ \Comment{Increase Penalty Strength}
\EndFor
\end{algorithmic}
\end{algorithm}

\begin{algorithm}[h!]\label{Algo3}
\caption{Multi-step NeuralODE - Formulation 3}\label{alg:cap}
\begin{algorithmic}
\While{ $j<$  n\_batches}
    \State get\_batch($t, \textbf{q}$, ..) \Comment{Get a new batch}
    \State Initailize $\boldsymbol{q_k}$ \Comment{Initialize discontinuities for the new batch}
    \State Params $\gets$($\boldsymbol{\Theta}$, $\boldsymbol{q_k}$) \Comment{Append neural network parameters and $\boldsymbol{q_k}$ as set of trainable parameters}
    \For{$\mu$ in  \{$\mu_i$\}} 
    \While{$k<$ num\_iters}
    \State ($\nabla \mathcal{L}_{\boldsymbol{\Theta}}, \nabla \mathcal{L}_{\boldsymbol{q_k}}$) = compute\_gradients($\mathcal{L}$,Params,$\mu_i$) \Comment{Compute gradients using Adjoint or autograd}
    \State Params $\gets$ optimizer.step(Params,$\nabla \mathcal{L}_{\boldsymbol{\Theta}}, \nabla \mathcal{L}_{\boldsymbol{q_k}}$) \Comment{Update the trainable parameters}
    \EndWhile
    \State update $\mu$ \Comment{Increase Penalty Strength}
\EndFor
\EndWhile
\end{algorithmic}
\end{algorithm}

\subsection{Scheduling Penalty Strength}\label{App ablation}
\begin{figure}[h]
    \centering
    \includegraphics[width=0.98\linewidth]{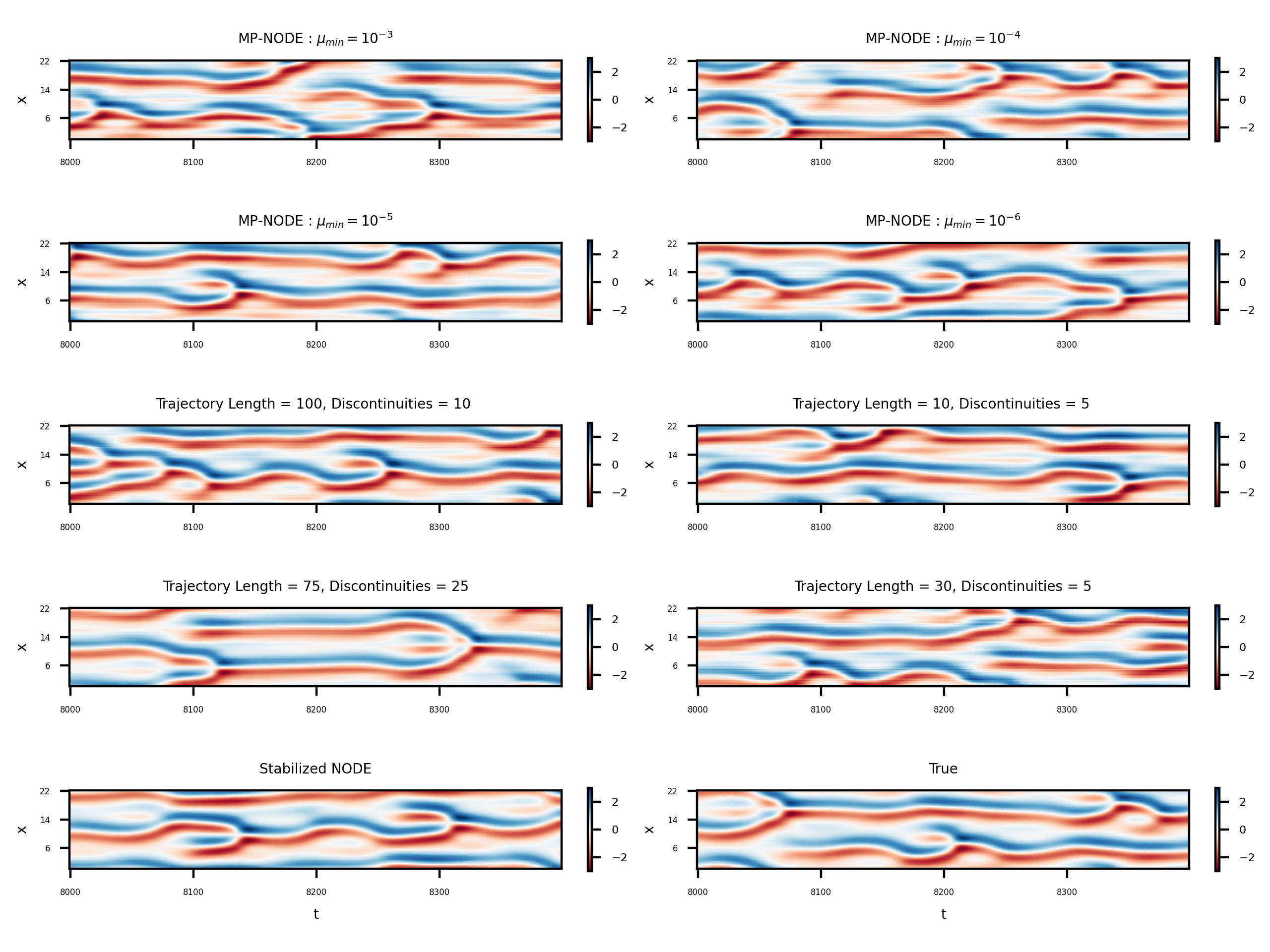}
    \caption{Model predictions for 1D KS system using varying hyperparameters.}
    \label{fig:mu ablation}
\end{figure}

The penalty strength $\mu$ in Equation \ref{MP loss equation} is an important hyper-parameter of the model. We use the Kuramoto-Shivashinsky(KS) equation(Section \ref{KS_problem}) to test the sensitivity the model performance with varying penalty strength. Figure \ref{fig:mu ablation} shows the predictions for 1D KS system using varying penalty strength $\mu$. The starting penalty strength for the optimization is given by $\mu_{min}$. It is increased by a factor of 10 every time the optimizer reaches a plateau (chosen as 10000 steps empirically). We also vary the total length of trajectory for optimization and the number of discontinuity points within the trajectory. These hyperparametrs are chosen based on their uniqueness to the MP optimization. All other hyper-parameters are consistent across all the models. 
We also compared the Kullback Leibler(KL) divergence for the joint probability distribution function (PDF) between the first and second derivative of the state for all the models and compared it to stabilized NODE\cite{linot2023stabilized} for reference. The KL divergence is computed using
\begin{equation}D_{KL}(\tilde{P}||P)=\int\limits_{-\infty}^{\infty}\int\limits_{-\infty}^{\infty}\tilde{P}(q_{x},q_{xx})\mathrm{ln}\frac{\tilde{P}(q_{x},q_{xx})}{P(q_{x},q_{xx})} dq_{x}dq_{xx}\end{equation}
where $\tilde{P}$ and $P$ are the joint PDFs for the model and ground truth respectively. It can be observed from Figure \ref{fig:mu ablation} that all the models perform decently well. However Table \ref{table_ablation} shows that $\mu_{min} = 10^{-4}$ and trajectory length of 75 timesteps with 25 discontinuity points in between performs best considering the KL divergences. However, it is problem dependent and need to be tuned for each problem. Extensive ensemble ablation studies for 2D turbulence and ERA5 is left for future research.

\begin{table}[h]
\begin{center}
\begin{tabular}{|c|c|c|c|c|}
\hline
    \textbf{Model Name} & $\mathbf{\mu_{min}}$ & \textbf{Trajectory Length} & \textbf{Discontinuities} & \textbf{KL div} \\\hline\\[-1em]
         MP-NODE 1 & $10^{-3}$&  60  & 5 &   0.07778  \\\hline\\[-1em]
         MP-NODE 2 & $10^{-4}$ & 60 & 5  &    0.06476             \\ \hline \\ [-1em]
         MP-NODE 3 & $10^{-5}$ & 60  & 5 &     0.08973            \\ \hline \\ [-1em]
         MP-NODE 4 & $10^{-6}$ & 60  & 5 &     0.13854            \\ \hline \\ [-1em]
         MP-NODE 5 & $10^{-4}$&  30  & 5 &   0.10639              \\ \hline \\ [-1em]
         MP-NODE 6 & $10^{-4}$ & 10 & 5  &    0.37875             \\ \hline \\ [-1em]
         MP-NODE 7 & $10^{-4}$ & 75  & 25 &     0.02915       \\ \hline \\ [-1em]
         MP-NODE 8 & $10^{-4}$ & 100  & 10 &     0.08691         \\ \hhline{|=|=|=|=|=|} \\ [-1em]
         Stab-NODE   & NA & NA  &NA &     0.04493         \\ \hline \\ [-1em]
         NODE & NA & NA & NA &  0.77327 \\ \hline
\end{tabular}
\end{center}
\caption{Table comparing the KL divergences of joint probability distribution function(PDF) between the first and second derivative of different models for predictions of Kuramoto-Shivashinsky equation. Stab-NODE and NODE models are also included as references to compare.}
\label{table_ablation}
\end{table}

\end{document}